\documentclass[10pt,journal,compsoc]{IEEEtran}
\ifCLASSOPTIONcompsoc
  \usepackage[nocompress]{cite}
\else
  \usepackage{cite}
\fi
\ifCLASSINFOpdf
\else
\fi

\usepackage{hyperref}       
\usepackage{url}            
\usepackage{booktabs}       
\usepackage{nicefrac}       
\usepackage{microtype}      
\usepackage{subfigure}
\usepackage{makecell}

\usepackage{enumitem}
\usepackage{amsmath}
\usepackage{enumerate}
\usepackage{xcolor}
\usepackage{bm}
\usepackage{graphicx}
\usepackage{color}
\usepackage{chngpage}
\usepackage{import}
\usepackage{amssymb,amsfonts}
\usepackage{textcomp}
\usepackage{flushend}
\usepackage{hhline}
\usepackage{multirow}
\usepackage{array}
\usepackage{gensymb}
\usepackage{nameref}
\usepackage{tabularx}
\usepackage[normalem]{ulem}
\usepackage{algorithm}
\usepackage{algpseudocode}
 \usepackage{hyperref}
\makeatletter
\algrenewcommand\ALG@beginalgorithmic{\small}
\algrenewcommand\algorithmiccomment[2][\small]{{#1\hfill\(\triangleright\) #2}}
\makeatother
\usepackage{enumitem}
\newenvironment{tight_itemize}{
\begin{itemize}[leftmargin=20pt]
  \setlength{\topsep}{0pt}
  \setlength{\itemsep}{0pt}
  \setlength{\parskip}{0pt}
  \setlength{\parsep}{0pt}
}{\end{itemize}}
\newcommand{\tabincell}[2]{\begin{tabular}{@{}#1@{}}#2\end{tabular}}

\DeclareRobustCommand\nobreakspace{\leavevmode\nobreak\ }
\newcolumntype{"}{!{\vrule width 1pt}}
\newcolumntype{L}[1]{>{\raggedright\let\newline\\\arraybackslash\hspace{0pt}}m{#1}}
\newcolumntype{C}[1]{>{\centering\let\newline\\\arraybackslash\hspace{0pt}}m{#1}}
\newcolumntype{R}[1]{>{\raggedleft\let\newline\\\arraybackslash\hspace{0pt}}m{#1}}

\begin{document}
\title{WebFace260M: A Benchmark for Million-Scale Deep Face Recognition}

\author{\normalsize{
Zheng Zhu, Guan Huang, Jiankang Deng, Yun Ye, Junjie Huang, \\ Xinze Chen, Jiagang Zhu, Tian Yang, Dalong Du, Jiwen Lu,  Jie Zhou

\IEEEcompsocitemizethanks{\IEEEcompsocthanksitem Zheng Zhu, Jiwen Lu, Jie Zhou are with Tsinghua University, Beijing, China. E-mail: \href{mailto:zhengzhu@tsinghua.edu.cn}{zhengzhu@tsinghua.edu.cn}
\IEEEcompsocthanksitem Guan Huang, Yun Ye, Junjie Huang, Xinze Chen, Jiagang Zhu, Tian Yang, Dalong Du are with XForwardAI, Beijing, China.
\IEEEcompsocthanksitem Jiankang Deng is with Imperial College London, London, UK.
\IEEEcompsocthanksitem Corresponding author: Jiwen Lu. E-mail: \href{mailto:lujiwen@tsinghua.edu.cn}{lujiwen@tsinghua.edu.cn}.
}
}
}

\markboth{IEEE TRANSACTIONS ON Pattern Analysis and Machine Intelligence, VOL. XX, NO. XX, XXX 2021}%
{Shell \MakeLowercase{\textit{et al.}}: WebFace260M: A Benchmark for Million-Scale Deep Face Recognition}

\IEEEtitleabstractindextext{%
\begin{abstract}
Face benchmarks empower the research community to train and evaluate high-performance face recognition systems. In this paper, we contribute a new million-scale recognition benchmark, containing uncurated 4M identities/260M faces (WebFace260M) and cleaned 2M identities/42M faces (WebFace42M) training data, as well as an elaborately designed time-constrained evaluation protocol. Firstly, we collect 4M name lists and download 260M faces from the Internet. Then, a Cleaning Automatically utilizing Self-Training (CAST) pipeline is devised to purify the tremendous WebFace260M, which is efficient and scalable. To the best of our knowledge, the cleaned WebFace42M is the largest public face recognition training set and we expect to close the data gap between academia and industry. Referring to practical deployments, Face Recognition Under Inference Time conStraint (FRUITS) protocol and a new test set with rich attributes are constructed. Besides, we gather a large-scale masked face sub-set for biometrics assessment under COVID-19. For a comprehensive evaluation of face matchers, three recognition tasks are performed under standard, masked and unbiased settings, respectively. Equipped with this benchmark, we delve into million-scale face recognition problems. A distributed framework is developed to train face recognition models efficiently without tampering with the performance. Enabled by WebFace42M, we reduce 40\% failure rate on the challenging IJB-C set and rank 3rd among 430 entries on NIST-FRVT. Even 10\% data (WebFace4M) shows superior performance compared with the public training sets. Furthermore, comprehensive baselines are established under the FRUITS-100/500/1000 milliseconds protocols. The proposed benchmark shows enormous potential on standard, masked and unbiased face recognition scenarios.
Our WebFace260M website is \url{https://www.face-benchmark.org}.
\end{abstract}


}

\maketitle

\IEEEdisplaynontitleabstractindextext

\IEEEpeerreviewmaketitle

\section{Introduction}
\IEEEPARstart{R}{ecognizing} faces in the wild has achieved remarkable success due to the boom of neural networks.
The key engine of recent face recognition consists of network architecture evolution~\cite{AlexNet,GoogleNet,VGGNet,ResNet,MobileNet,wu2018light,Deepid3}, a variety of loss functions~\cite{DeepFace,FaceNet,DeepID,DeepID2,wen2016discriminative,deng2017marginal,NormFace,A-SoftMax,CosFace,AM-SoftMax,ArcFace,Curricularface,Groupface,BroadFace}, and growing face benchmarks~\cite{LFW,AgeDB,CFP,CALFW,CPLFW,MegaFace,IJB-C,CASIA-WebFace,VGGFace,VGGFace2,UMDFaces,MF2,MS1M,IMDB-Face}.
Even though growing efforts have been devoted to investigating sophisticated networks and losses, academia is restricted by limited training sets and nearly saturated test protocols.

\begin{table*}[!t]
\caption{Training sets for deep face recognition. The curated WebFace42M is the largest public training data in terms of both \# identities and \# images.}
\begin{center}{\scalebox{1.0}{
\begin{tabular}{l|c|c|c|c|c|c|c}
\hline
Dataset & \# Identities & \# Images & Images/ID  & Cleaning  & \# Attributes & Availability & Publications\\ \hline
\hline
CASIA-WebFace~\cite{CASIA-WebFace} & $10$ K & $0.5$ M & 47 & Automatic & - & Public & Arxiv 2014\\
CelebFaces~\cite{DeepID} & $10$ K & $0.2$ M & 20 & Manual & 40  & Public & ICCV 2015\\
VGGFace~\cite{VGGFace} & $2$ K & $2.6$ M & 1,000 & Semi-automatic & - & Public & BMVC 2015\\
MS1M~\cite{MS1M} &$0.1$ M & $10$ M & 100 & No & -& Public & ECCV 2016\\
MegaFace2~\cite{MF2} & $0.6$ M & $4.7$ M & 7 & Automatic & -& Public & CVPR 2017\\
MS1M-IBUG~\cite{deng2017marginal} &$85$ K & $3.8$ M & 45 & Semi-automatic & -& Public & CVPRW 2017\\
UMDFaces~\cite{UMDFaces} & $8$ K & $0.3$ M & 45 & Semi-automatic & 4 & Public & IJCB 2017\\
IMDB-Face~\cite{IMDB-Face} & $59$ K & $1.7$ M & 29 & Manual &  - & Public & ECCV 2018\\

VGGFace2~\cite{VGGFace2} & $9$ K & $3.3$ M & 363 & Semi-automatic & 11 & Public & FG 2018\\

MS1MV2~\cite{ArcFace} & $85$ K & $5.8$ M & 68 & Semi-automatic & -& Public & CVPR 2019\\
MS1M-Glint~\cite{glintweb} & $87$ K & $3.9$ M  & 44 & Semi-automatic & -& Public & -\\
\hline
\textcolor[RGB]{128,128,128}{Facebook~\cite{DeepFace}} & \textcolor[RGB]{128,128,128}{$4$ K}  & \textcolor[RGB]{128,128,128}{$4.4$ M} & \textcolor[RGB]{128,128,128}{1,100} & \textcolor[RGB]{128,128,128}{-} & \textcolor[RGB]{128,128,128}{-}& \textcolor[RGB]{128,128,128}{Private} &\textcolor[RGB]{128,128,128}{CVPR 2014}\\
\textcolor[RGB]{128,128,128}{Facebook~\cite{taigman2015web}} & \textcolor[RGB]{128,128,128}{$10$ M} & \textcolor[RGB]{128,128,128}{$500$ M} & \textcolor[RGB]{128,128,128}{50} & \textcolor[RGB]{128,128,128}{-} & \textcolor[RGB]{128,128,128}{-}& \textcolor[RGB]{128,128,128}{Private} &\textcolor[RGB]{128,128,128}{CVPR 2015}\\
\textcolor[RGB]{128,128,128}{Google~\cite{FaceNet}} & \textcolor[RGB]{128,128,128}{$8$ M} & \textcolor[RGB]{128,128,128}{$200$ M} & \textcolor[RGB]{128,128,128}{25} & \textcolor[RGB]{128,128,128}{-} & \textcolor[RGB]{128,128,128}{-}& \textcolor[RGB]{128,128,128}{Private} & \textcolor[RGB]{128,128,128}{CVPR 2015}\\
\textcolor[RGB]{128,128,128}{MillionCelebs~\cite{MillionCelebs}} & \textcolor[RGB]{128,128,128}{$0.6$ M} & \textcolor[RGB]{128,128,128}{$18.8$ M} & \textcolor[RGB]{128,128,128}{30} & \textcolor[RGB]{128,128,128}{Automatic} & \textcolor[RGB]{128,128,128}{-}& \textcolor[RGB]{128,128,128}{Private} & \textcolor[RGB]{128,128,128}{CVPR 2020}\\ \hline
\textbf{WebFace260M}  & \textbf{4 M}  & \textbf{260M}  & \textbf{65} & No & -& Public & - \\
\textbf{WebFace42M} & \textbf{2 M}  & \textbf{42M}  & \textbf{21} & Automatic & 7 & Public & - \\ \hline
\end{tabular}}}
\end{center}
\label{table:training_set}
\end{table*}

\begin{figure}[t]
\centering
{
\includegraphics[width=0.94\linewidth]{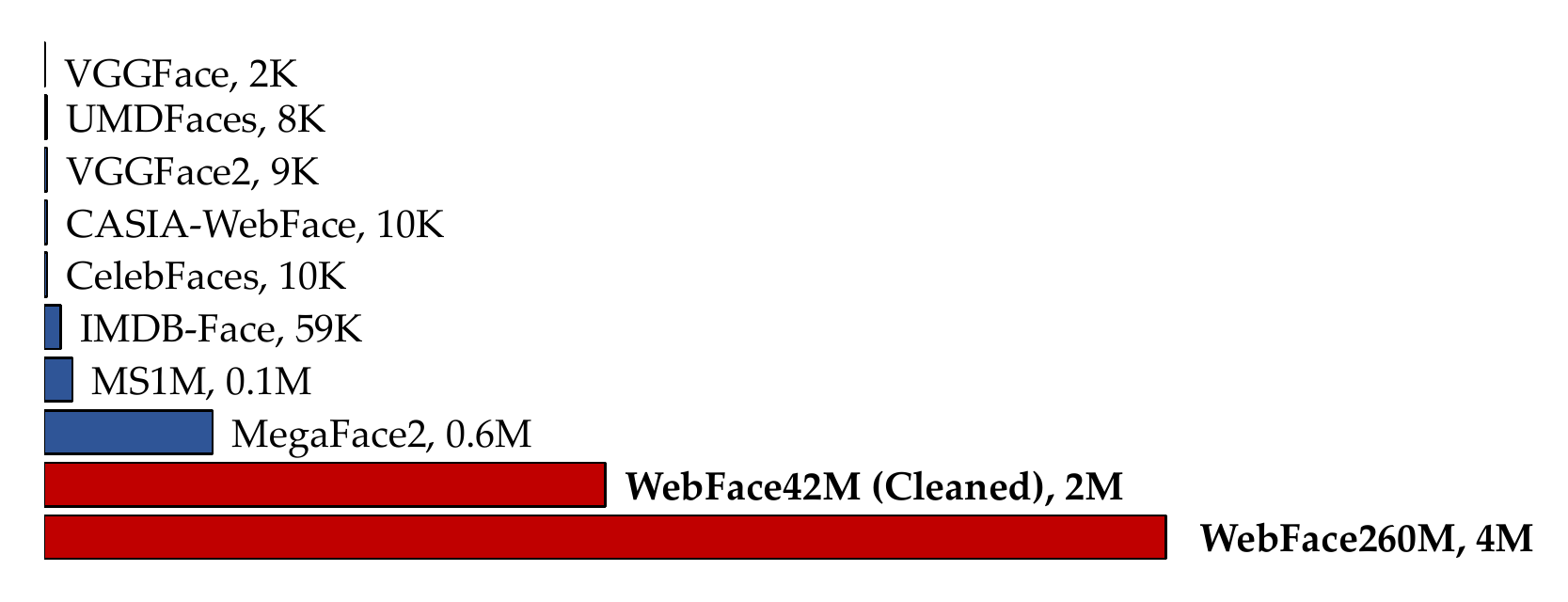}}
{
\includegraphics[width=0.94\linewidth]{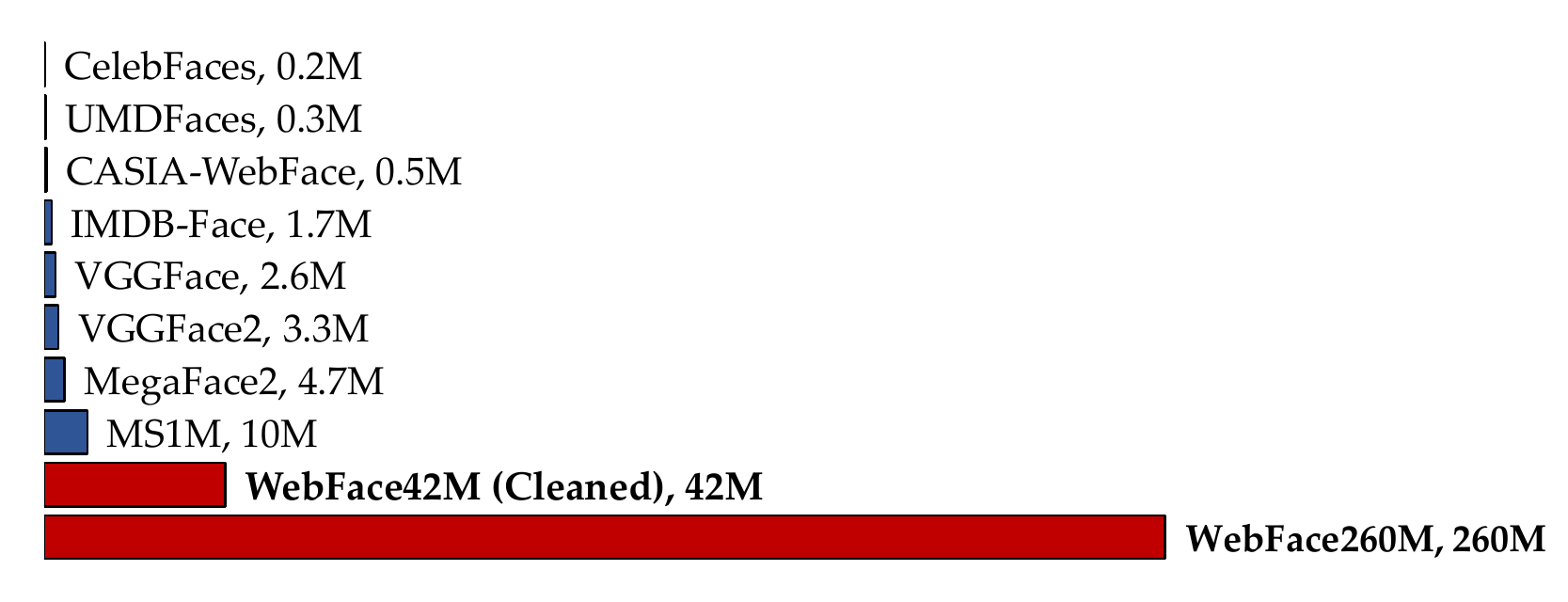}}
\caption{Comparisons of \# identities and \# faces for our WebFace data and public training sets.}
\label{fig:training_set}
\vspace{-4mm}
\end{figure}

 As shown in Table~\ref{table:training_set} and Figure~\ref{fig:training_set}, the largest public training sets in terms of identities and faces are MegaFace2~\cite{MF2} and MS1M~\cite{MS1M}, respectively. MegaFace2 contains 4.7M faces of 672K subjects collected from Flickr~\cite{Flickr}. MS1M consists of 10M faces of 100K celebrities but the noise rate is around 50\%~\cite{IMDB-Face}. In contrast, companies from industry can access much larger private data to train face recognition models: Google utilizes 200M images of 8M identities to train FaceNet~\cite{FaceNet}, and Facebook~\cite{taigman2015web} performs training by 500M faces of 10M identities. This data gap hinders researchers from pushing the frontiers of deep face recognition. The main obstacles for tremendous training data lie in large-scale identity collection, effective and scalable cleaning, and efficient training. For example, IMDB-Face~\cite{IMDB-Face} takes 50 annotators to work continuously for one month to obtain 59K identities and 1.7M images, which is labor-intensive and non-scalable.

On the other hand,  test sets and protocols play an essential role in analyzing face recognition performance.
Popular evaluations including LFW families~\cite{LFW,CALFW,CPLFW}, CFP~\cite{CFP}, AgeDB~\cite{AgeDB}, RFW~\cite{RFW}, MegaFace~\cite{MegaFace}, and IJB-C~\cite{IJB-C} mainly target the pursuit of accuracy, which have been almost saturated recently. In real-world application scenarios, face recognition is always restricted by inference time, such as unlocking mobile telephones with a smooth user experience. Lightweight Face Recognition (LFR) Challenge~\cite{LFR} takes a step toward this goal by constraining model size and FLOPs, but actual inference time can vary quite a bit for different networks.
Besides, LFR neglects the time cost of face detection and alignment. To the best of our knowledge, NIST-FRVT~\cite{FRVT} is the only time-constrained face recognition protocol. However, the strict submission policy (no more than one submission every four calendar months) hinders researchers from freely evaluating their algorithms.

To address the above problems, this paper constructs a new ultra-large-scale face benchmark consisting of 4M identities/260M faces (WebFace260M) as well as a time-constrained assessment protocol. Firstly, a name list of 4M celebrities is gathered and 260M images are downloaded utilizing a search engine. Then, we perform Cleaning Automatically by Self-Training (CAST) pipeline, which is scalable and does not need any human intervention. The proposed CAST procedure results in high-quality 2M identities and 42M faces (WebFace42M).
Meanwhile, rich attributes are provided for further analyzing the statistics of WebFace42M.
Referring to various real-world applications, we design the Face Recognition Under Inference Time conStraint (FRUITS) protocol, enabling academia to test deep face matchers comprehensively. Specifically, the FRUITS includes three time limit tracks: 100, 500, and 1000 milliseconds, which intend to evaluate deployments on mobile devices, local devices, and clouds, respectively.
Since public evaluations are almost saturated ~\cite{LFW, AgeDB, CFP} and may contain noise~\cite{MegaFace,IJB-C}, we manually construct a new test set with rich attributes to enable FRUITS.
Considering the COVID-19 coronavirus epidemic \cite{eikenberry2020mask,kwon2021association} and reported biased face recognition deployments \cite{RFW,gong2020jointly,wang2019mitigate}, three evaluation tasks are performed: Standard Face Recognition (SFR), Masked Face Recognition (MFR), and Unbiased Face Recognition (UFR). For MFR, we collect a large-scale masked face test sub-set.


Based on the proposed ultra-large-scale benchmark, we delve into million-scale deep face recognition problems. With such data size, a distributed training framework is developed for efficient optimization, which could perform at a nearly linear acceleration without performance drops. Accuracy on the public and proposed test sets indicates that our training data is indispensable for pushing the frontiers of deep face recognition: WebFace42M achieves 97.70\% TAR@FAR=1e-4 on the challenging IJB-C \cite{IJB-C} under standard ResNet-100 configurations, reducing near 40\% relative error rate compared with state of the art. 10\% of our data (WebFace4M) also obtains superior performance than similar-sized MS1M families \cite{deng2017marginal,ArcFace,glintweb} and MegaFace2 \cite{MF2}.
On the proposed test set, a similar conclusion can be drawn.
For SFR, WebFace42M decreases FNMR@FMR=1e-5 from 9.88\% (with MS1MV2) to 2.98\% under the same settings, which reduces the error rate by 3 times.
For MFR, error rates (FNMR@FMR=1e-5) of WebFace42M and MS1MV2 are 42.97\% and 69.56\%, respectively. For UFR, considering the tremendous scale of the WebFace data, it provides more room for data balancing and fairness exploration. Furthermore, we participate in the NIST-FRVT~\cite{FRVT} and rank 3rd among 430 entries based on WebFace42M. At last, we discuss privacy and bias issues in this benchmark.

For baseline comparisons, comprehensive deep face recognition systems are evaluated under FRUITS-100/500/1000 milliseconds protocols.
For SFR, different settings of
face detection/alignment and feature extraction are explored, covering
MobileNet \cite{MobileNet,chen2018mobilefacenets}, EfficientNet \cite{tan2019efficientnet}, AttentionNet \cite{AttentionNet}, ResNet \cite{ResNet}, SENet \cite{SENet}, ResNeXt \cite{ResNeXt} and RegNet families \cite{RegNet}.
For MFR, the influence of mask augmentation in WebFace training data is studied and a strong baseline is established for this difficult problem.
In order to investigate the bias in current face recognition systems, we re-sample the WebFace data to obtain an attribute-balanced sub-set and evaluate its influence on recognition fairness.
With this new face benchmark, we hope to close the data gap between academia and industry, and facilitate the time-constrained recognition assessment for real-world applications.

The main contributions of this work can be summarized as follows:
\begin{tight_itemize}
\item An ultra-large-scale face recognition dataset is constructed for the research community towards closing the data gap behind the industry.
The proposed WebFace260M consists of 4M identities and 260M faces, which provides an excellent resource for million-class deep face cleaning and recognition as shown in Table~\ref{table:training_set} and Figure~\ref{fig:training_set}.

\item We contribute the training set WebFace42M which sets new state of the art on challenging IJB-C and ranks 3rd on NIST-FRVT. This cleaned data is automatically purified from WebFace260M by a scalable and effective self-training pipeline. To the best of our knowledge, the resulting WebFace42M is the largest public face recognition training resource.
\item The FRUITS protocol as well as a test set with rich attributes are constructed to facilitate the evaluation of real-world applications. Meanwhile, we collect a large-scale masked face set. A series of time-constrained tasks are designed referring to different deployment scenarios, including standard, masked, and unbiased face recognition.
\item Based on the new benchmark, we perform extensive million-scale face recognition experiments. Enabled by distributed training framework, comprehensive baselines are established on public and our test sets under the FRUITS protocol. The results indicate substantial insights on three different recognition settings.
\end{tight_itemize}

This paper is built upon our conference work \cite{WebFace260M} and significantly extended in several aspects. Firstly, we provide comprehensive visualization results to illustrate the face benchmark.
This gives a deeper insight into the diversity and challenges of our training/test data.
Secondly, the scale of our test set is significantly increased, making it more challenging for recognition evaluation. Besides, we collect a highly-curated masked sub-set, which contains 862 subjects with real-world masks. Lastly, for evaluation and experiment, standard face recognition is extended to masked and unbiased settings. Corresponding mask augmentation and attribute-balanced baselines are also established. In addition, we present a detailed literature review for
deep face recognition and benchmarks. Privacy and bias issues in WebFace260M benchmark are also discussed.

Since the preliminary version of this work was published, we have received dataset access applications \footnote{\url{https://www.face-benchmark.org/download.html}} from near 400 research groups. Based on the WebFace260M benchmark, we organize the Face Bio-metrics under COVID Workshop and Masked Face Recognition Challenge \cite{zhu2021masked,deng2021masked} in ICCV 2021. More than 80 teams have participated in WebFace260M Track under FRUITS protocol and submitted more than 1,000 solutions \footnote{\url{https://competitions.codalab.org/competitions/32478\#results}}. In InsightFace Unconstrained Track, all top-3 teams from academia and industry adopt WebFace260M database as their training sets.
These results suggest that our WebFace260M is not only an effective benchmark to pursue high-performance face recognition systems but also a meaningful step to reduce the data gap between research laboratories and companies.

\section{Related Works}

Face recognition has been extensively studied in the computer vision literature.
Recent years have witnessed a significant advance in both benchmarks and algorithms, including growing training data, evolutional evaluation sets, and loss function designs.
This section reviews representative progresses in academia and industry.

\subsection{Face Recognition Training Data}

A key aspect in developing face recognition systems is the training data used to learn discriminative face representations. Data collections are extremely important but usually overlooked. Even though some companies have internally labeled private face sets that scale to millions of images \cite{DeepFace} or even millions of subjects (Google \cite{FaceNet}and Facebook \cite{taigman2015web}), the situation is quite different for publicly available collections. As shown in Table~ \ref{table:training_set}, we give the detailed statistics of widely used training sets in the community, such as CASIA-WebFace~\cite{CASIA-WebFace}, VGGFace2~\cite{VGGFace2}, UMDFaces~\cite{UMDFaces}, MS1M~\cite{MS1M}, MegaFace2~\cite{MF2}, and IMDB-Face~\cite{IMDB-Face}.

CASIA-WebFace~\cite{CASIA-WebFace}, VGGFace2~\cite{VGGFace2} and UMDFaces~\cite{UMDFaces} consist of around 10K identities.
CASIA-WebFace~\cite{CASIA-WebFace} is collected by a semi-automatical method, which searches face images of celebrities from the Internet.
VGGFace2~\cite{VGGFace2} is an improved version of VGGFace~\cite{VGGFace} created in order to mitigate the deficiency of its predecessor. The subjects in VGGFace2 are collected from celebrities and famous people such as professors and politicians. Compared to its predecessor, VGGFace2 contains fewer images for each identity but covers a large range of poses, ages and races. To reduce label noise as much as possible, manual and automatic processes are employed. UMDFaces~\cite{UMDFaces} utilizes a mix of human annotators via Amazon Mechanical Turk (AMT) and pre-trained deep-based face analysis tools to build a face dataset that is much tougher than already available sets. 

MS1M~\cite{MS1M}, MegaFace2~\cite{MF2}, and IMDB-Face~\cite{IMDB-Face} include more identities than above-mentioned datasets. MS1M~\cite{MS1M} retrieves around 100 images for each identity by the Bing search engine \cite{bing_image} using the celebrity's name without any filtering. Therefore, the quality of MS1M is severely biased by label noises, duplicated images, and non-face images present in the set. All of these factors make MS1M hard to be used directly. MegaFace2~\cite{MF2} contains 672K subjects cleaned from Flickr. However, this dataset only collects 4.7M faces, which results in around 7 images per identity. IMDB-Face~\cite{IMDB-Face} claims to be the largest noise-controlled face collection, which contains 1.7M images of 59K celebrities by manual annotation. However, it took 50 annotators to work continuously for one month to clean the dataset, which demonstrates the difficulty of obtaining a large-scale clean dataset for face recognition.

\subsection{Face Recognition Evaluation}
\label{recognition_evaluation}

Most popular evaluation sets for face recognition target the pursuit of accuracy. CFP~\cite{CFP}, AgeDB~\cite{AgeDB}, CALFW~\cite{CALFW} and CPLFW~\cite{CPLFW} evaluate the verification accuracy under different intra-class variations (such as pose and age). MegaFace~\cite{MegaFace} and IJB-C~\cite{IJB-C} serve for both accuracies of large-scale face verification and identification. YTF~\cite{YTF} and IQIYI-Video~\cite{IQIYI2018} compare the accuracy of video-based verification. Different model-ensemble and post-processing \cite{PFE} could be adopted for higher performance under these protocols. However, face recognition in real-world application scenarios is always restricted by inference time, such as unlocking mobile phones with a smooth experience or processing multiple channels of surveillance videos on clouds.

Recently, the LFR Challenge \cite{LFR} takes a step toward this goal by constraining the FLOPs and model size of submissions. Since different neural network architectures can be quite different in terms of real inference time, this protocol is not a straightforward solution. Furthermore, it does not consider face detection and alignment, which are prerequisite components in most modern face recognition systems. To the best of our knowledge, NIST-FRVT~\cite{FRVT} is the only benchmark employing the time-constrained protocol. However, the strict submission policy (participants can only send one submission every four calendar months) hinders researchers from freely evaluating their algorithms.

 With global COVID-19 pandemic and reported biased systems deployment, up-to-date face recognition datasets focus on comparisons with masks and fairness consideration. RFW \cite{RFW} aims to evaluate the bias among 4 racial distributions. NIST-FRVT \cite{grother2019face} regularly evaluates the bias level of submitted algorithms caused by demographic effects. Due to the sudden outbreak of the coronavirus epidemic, there is yet no comprehensive real-world masked face recognition benchmark. Evaluations on simulated masked images \cite{FRVT-mask, negi2021deep} may result in questioned conclusions, and small-scale masked sets \cite{anwar2020masked,damer2021extended,IJCB-mask} can not comprehensively reflect the performance of algorithms.
Real-world masked test set RMFRD \cite{wang2020masked} consists of 525 identities and 5K masked faces, but there exist annotation noises.

\subsection{Deep Face Recognition}

The last decade has witnessed the advance of deep convolutional face recognition techniques.
A number of successful face recognition systems, such as DeepFace~\cite{DeepFace}, DeepID~\cite{DeepID,DeepID2,DeepID2+,Deepid3}, FaceNet~\cite{FaceNet} have achieved impressive performance on face verification and identification.
Most of the early works rely on metric-learning based losses~\cite{chopra2005learning,FaceNet,sohn2016improved}, and recent researches have switched to margin-based softmax losses due to their efficiency on the large-scale dataset.
SphereFace~\cite{A-SoftMax}, AM-softmax~\cite{AM-SoftMax}, CosFace~\cite{CosFace}, ArcFace~\cite{ArcFace} progressively improve the performance on various benchmarks to the newer level.
To further improve the margin-based softmax loss, recent works focus on the exploration of adaptive parameters \cite{zhang2019p2sgrad,zhang2019adacos,liu2019adaptiveface,liu2019fair}, inter-class regularization \cite{zhao2019regularface,duan2019uniformface}, sample mining \cite{wang2019mis,Curricularface}, learning acceleration \cite{zhang2018accelerated,an2020partial,li2021virtual,li2021dynamic,BroadFace}, etc.
There are also many complementary methods proposed to build better face recognition models by promoting desired properties of the produced face representations, such as robustness to noisy labels \cite{hu2019noise,Zhong2019Unequal,wang2019co,MillionCelebs,subcenter}, occlusions \cite{song2019occlusion,wang2020hierarchical,zheng2021learning} and low quality \cite{PFE,shi2020towards}, invariance to age \cite{wang2019decorrelated,zhao2020towards,huang2021age} and pose \cite{zhao2018towards,wang2021pseudo}, ability to mitigate racial bias \cite{RFW,wang2019mitigate,gong2021mitigating} and domain imbalance \cite{du2020semi,Groupface,cao2020domain,li2021dynamic}, to improve the fairness of representations \cite{liu2019fair,xu2021consistent}.


\section{WebFace260M and WebFace42M}

\subsection{Data Collection and Pre-processing}
\label{Data_Collection}
Knowledge graphs website Freebase~\cite{freebase} and well-curated website IMDB~\cite{imdb_website} provide excellent resources for collecting celebrity names. Furthermore, commercial search engines such as Google~\cite{google_image} and Bing \cite{bing_image} make it possible to collect images of a specific identity with ranked correlation. Our celebrity name list consists of two parts: the first one is borrowed from MS1M (1M, constructed from the Freebase) and the second one is collected from the IMDB database. There are near 4M celebrity names on the IMDB website, while we find some subjects have no public image from search engines. Therefore, only 3M celebrity names in IMDB are chosen for our benchmark. Based on the name list, celebrity faces are searched and downloaded via Google image search engine~\cite{google_image}. 200 images per identity are downloaded for the top 10\% subjects, while 100, 50, 25 images are reserved for the remaining 20\%, 30\%, 40\% subjects, respectively. Finally, we collect 4M identities and 265M images.

\begin{figure}
\small
\centering
\subfigure[Date of birth]{
\label{fig:allbirth}
\includegraphics[width=0.14\textwidth]{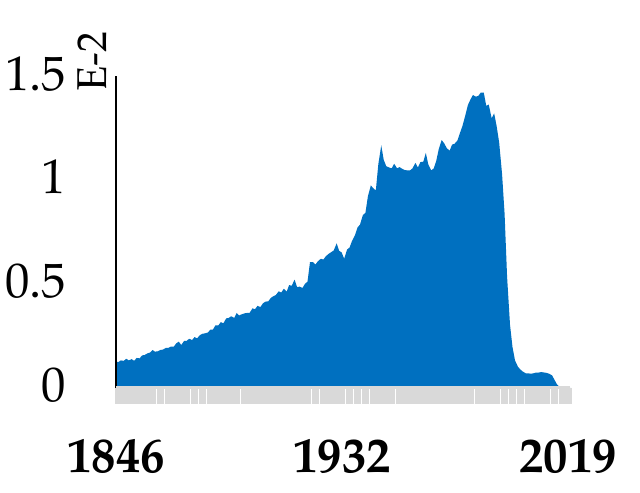}}
\subfigure[Nationality]{
\label{fig:allnationality}
\includegraphics[width=0.135\textwidth]{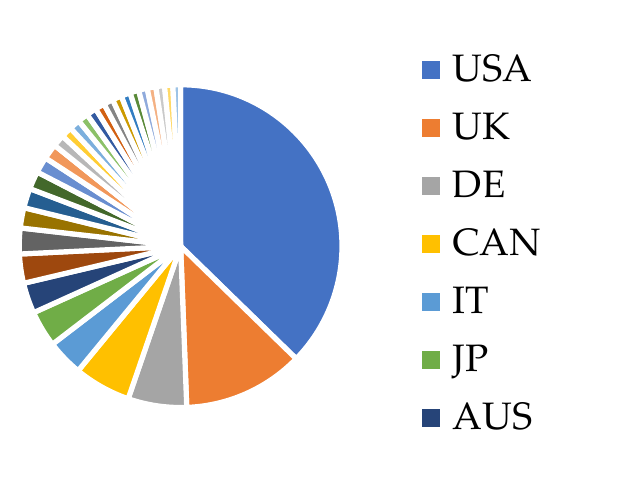}}
\subfigure[Profession]{
\label{fig:allprofession}
\includegraphics[width=0.15\textwidth]{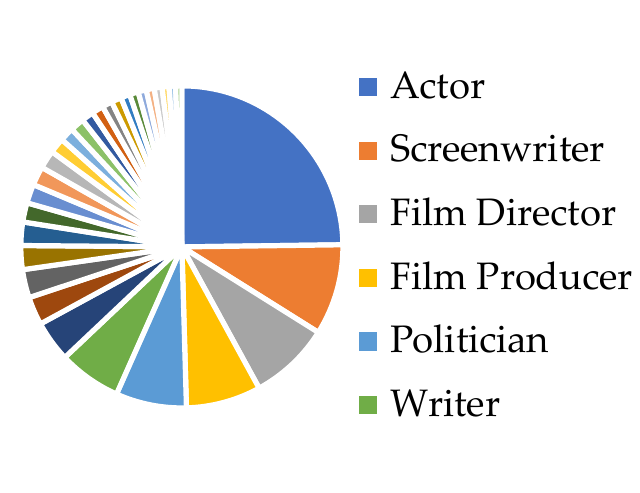}}
\subfigure[Pose]{
\label{fig:pose}
\includegraphics[width=0.14\textwidth]{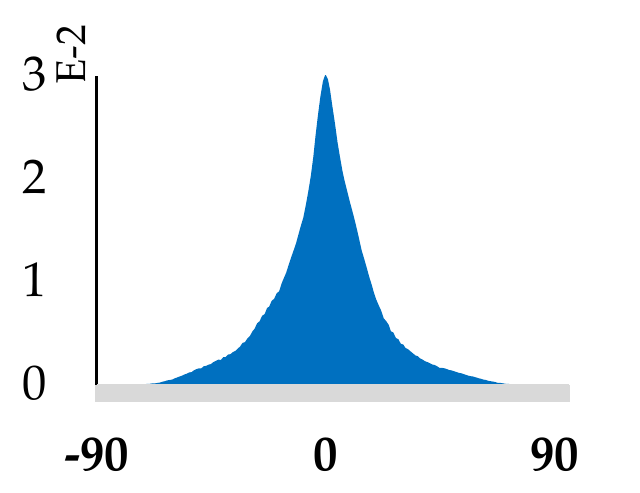}}
\subfigure[Age]{
\label{fig:age}
\includegraphics[width=0.14\textwidth]{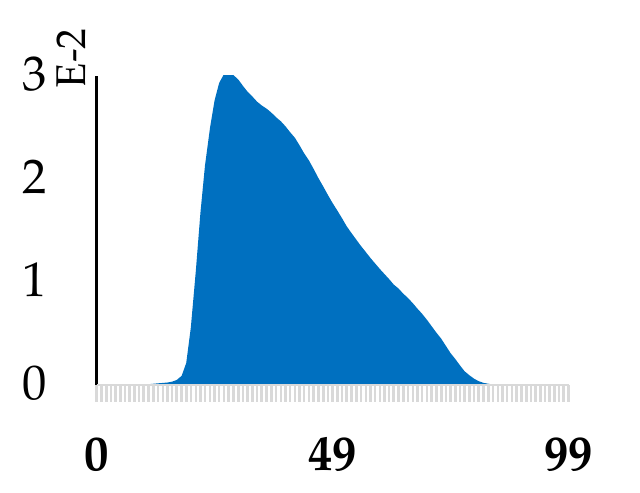}}
\subfigure[Race]{
\label{fig:race}
\includegraphics[width=0.15\textwidth]{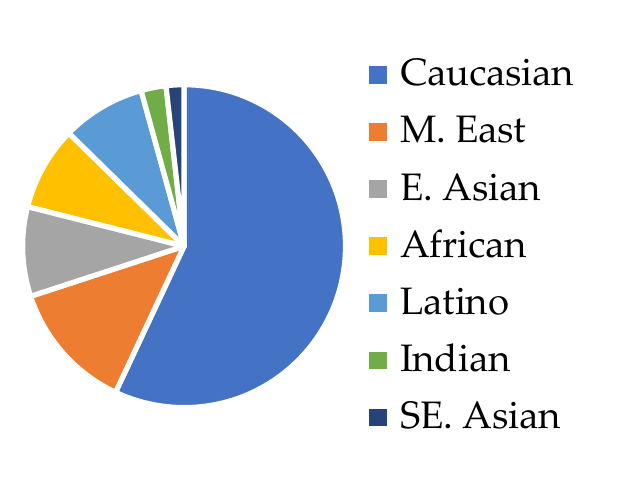}}
\caption{Birth date, nationality, profession distributions of WebFace260M,
and pose (yaw), age, race distributions of WebFace42M.}
\label{fig:face_attributes}
\end{figure}

In data pre-processing, faces are detected and aligned through five landmarks predicted by RetinaFace~\cite{RetinaFace}. Specifically, the threshold of detection score is set as 0.7 to filter the low confident faces. After pre-processing, there are 4M identities/260M faces (WebFace260M) shown as Table~\ref{table:training_set}. The statistics of WebFace260M are illustrated in Figure~\ref{fig:face_attributes} including date of birth, nationality and profession. Persons in WebFace260M come from more than 200 distinct countries/regions and more than 500 different professions with the date of birth back to 1846, which guarantees a great diversity in our training data.
\emph{During the construction of WebFace260M dataset, privacy and bias problems are our first concerns. Detailed discussion is available in Section~\ref{sec:discussion}}.


\begin{figure*}
\small
\centering
\subfigure[Celebrity \emph{Kalenna Harper}]{
\label{fig:Kalenna}
\includegraphics[width=0.47\linewidth]{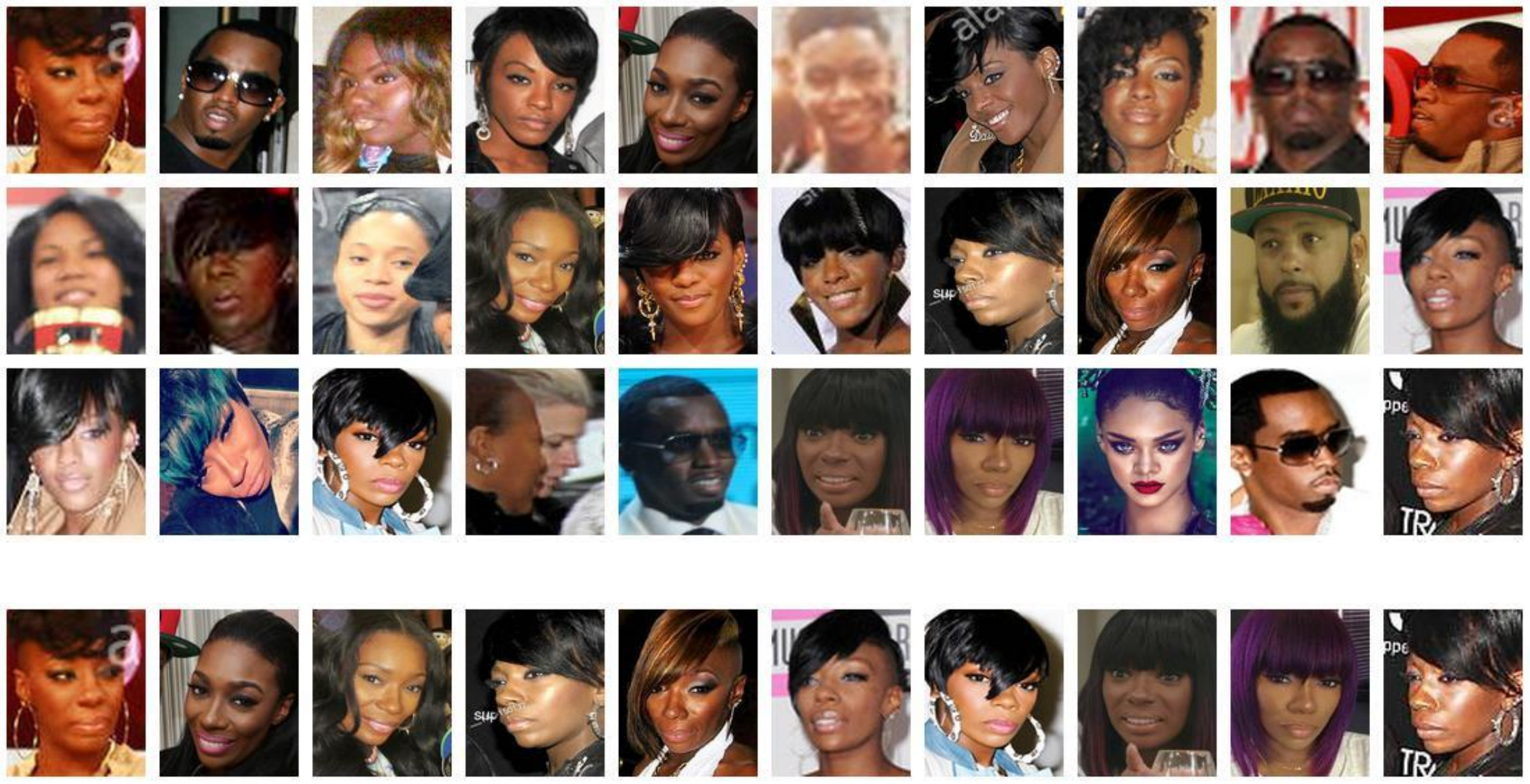}}
\subfigure[Celebrity \emph{Claire Ayer}]{
\label{fig:Claire}
\includegraphics[width=0.47\linewidth]{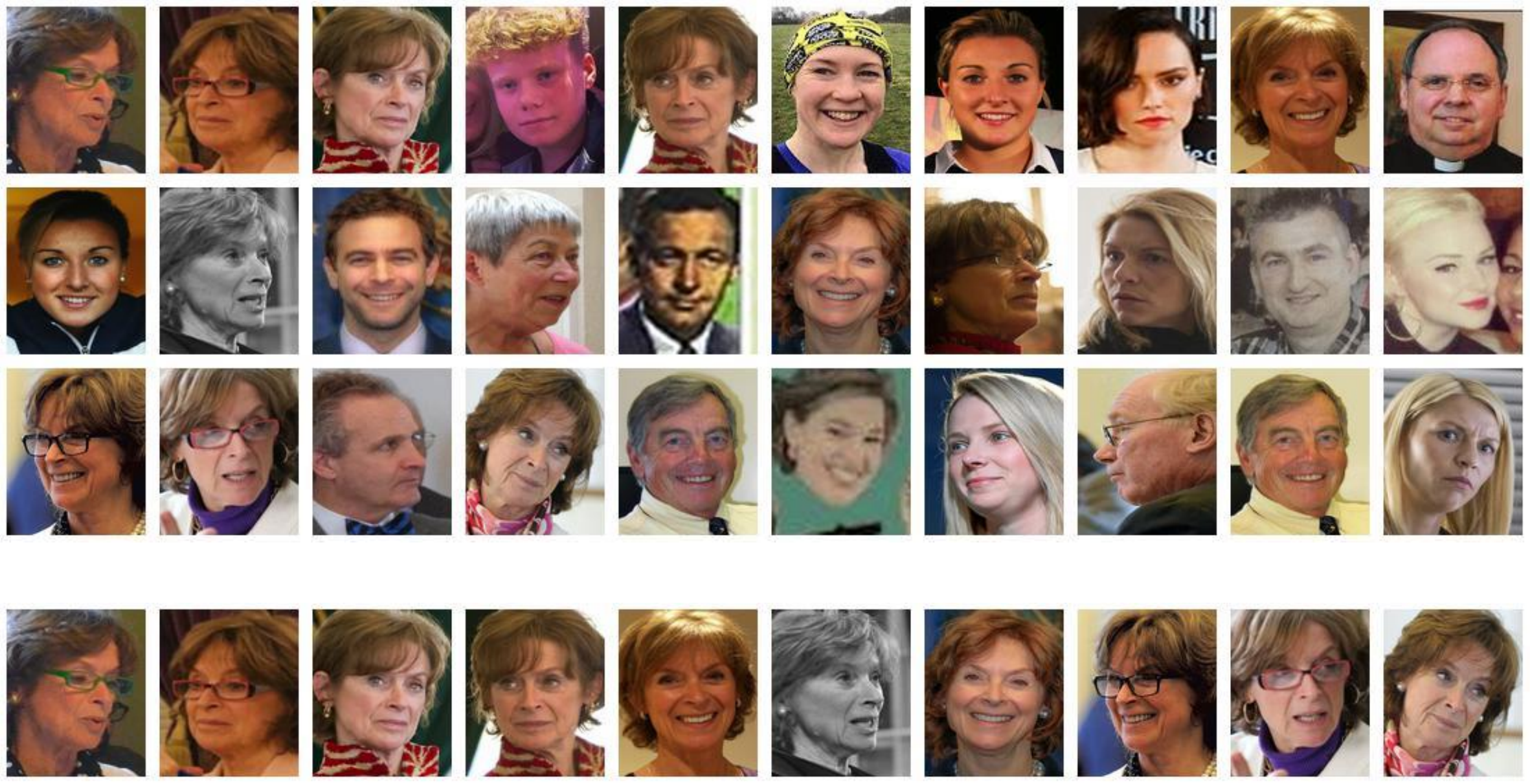}}
\subfigure[Celebrity \emph{Gregg Wallace}]{
\label{fig:Gregg}
\includegraphics[width=0.47\linewidth]{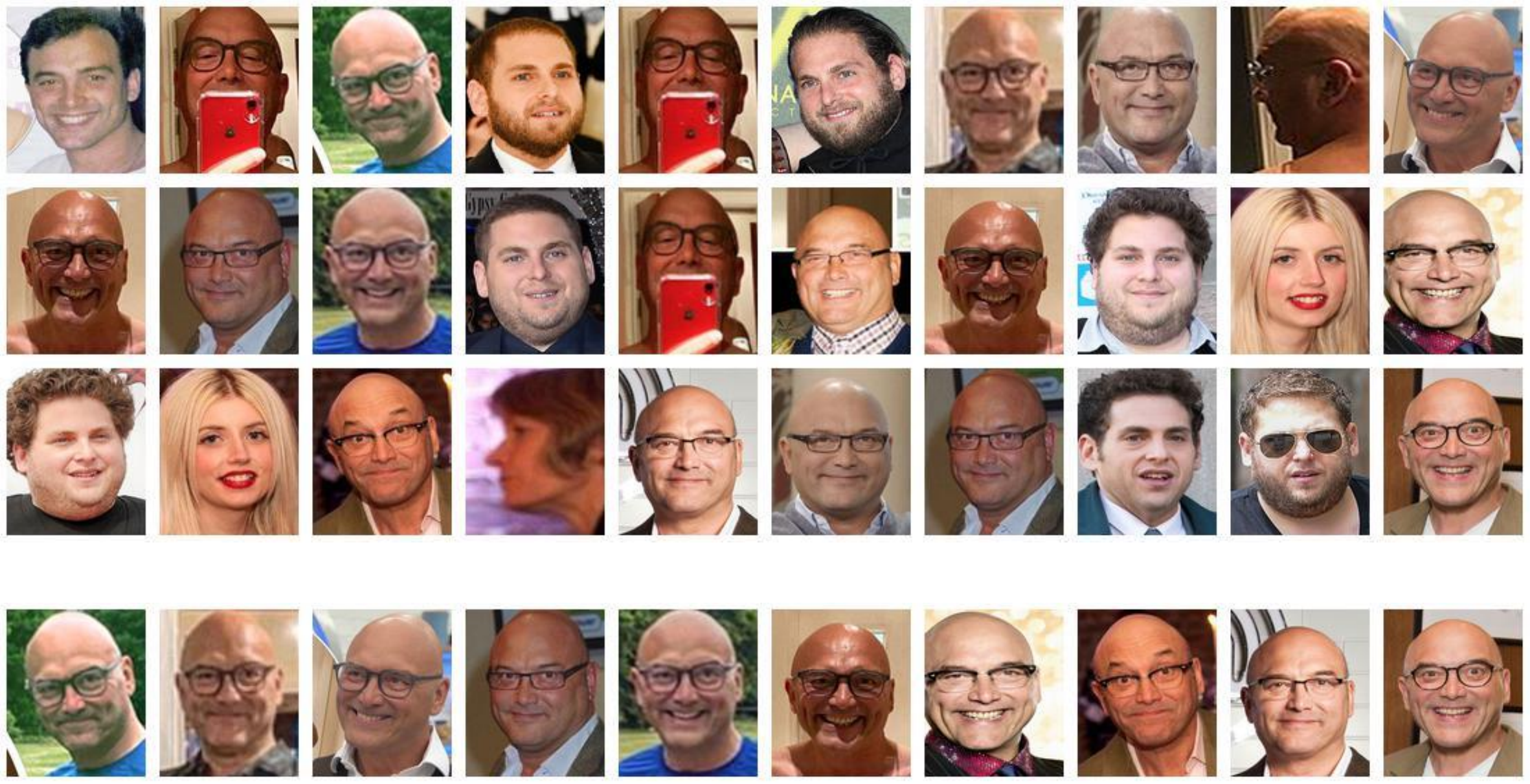}}
\subfigure[Celebrity \emph{Shaun Weiss}]{
\label{fig:Shaun}
\includegraphics[width=0.47\linewidth]{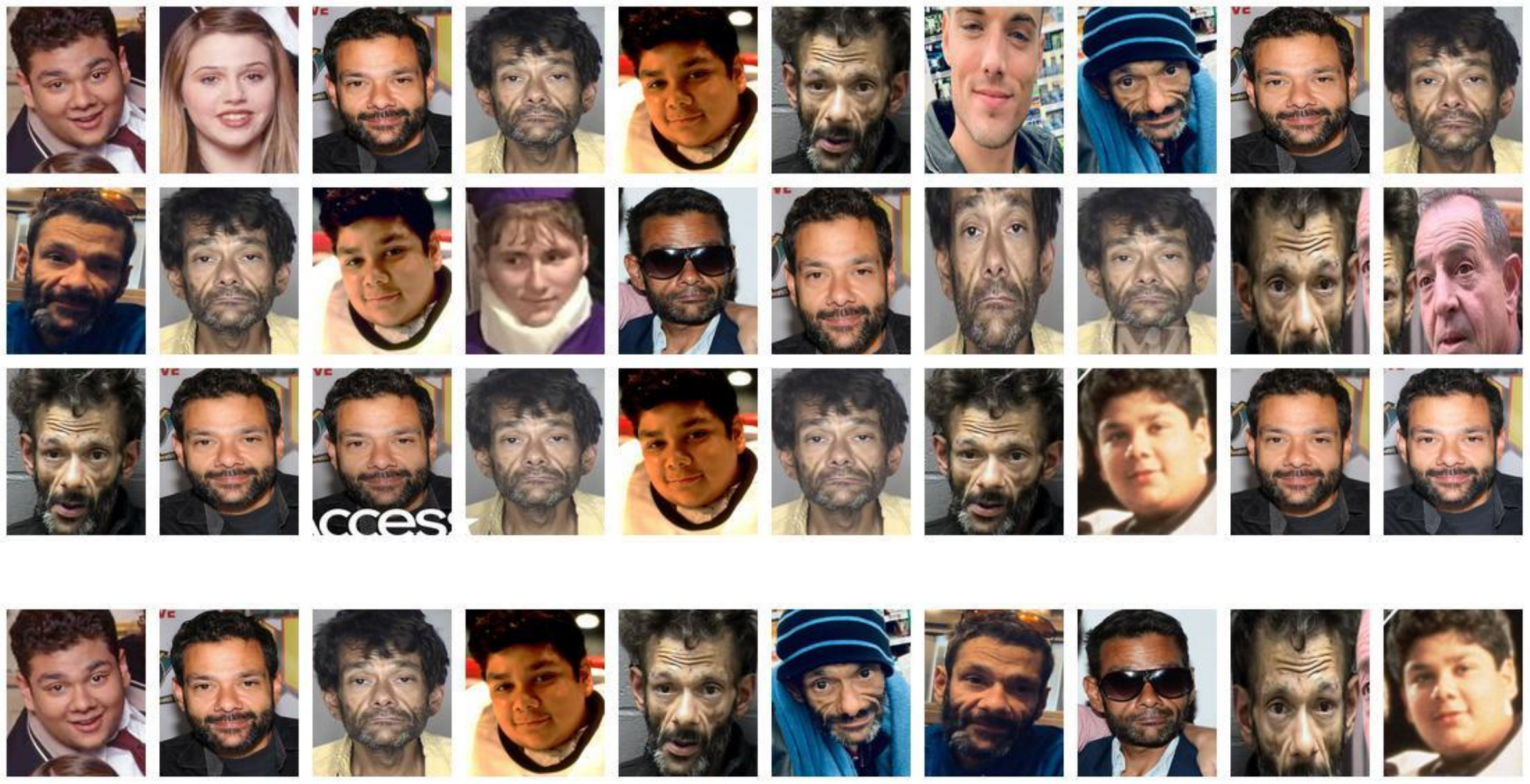}}

\caption{Visualization of the WebFace data. For each sub-figure, the top part is the randomly selected faces from WebFace260M, while the bottom part shows cleaned faces (also randomly selected) from WebFace42M. Loose cropped faces are shown.}
\label{fig:training_faces}
\vspace{-1mm}
\end{figure*}

\subsection{Cleaned WebFace42M}
We perform a CAST pipeline (Section~\ref{sec:cast}) to automatically clean the noisy WebFace260M and obtain a curated training set named WebFace42M, consisting of 42M faces of 2M subjects. Face number in each identity varies from 3 to more than 300, and the average face number is 21 per identity. As shown in Table~\ref{table:training_set} and Figure~\ref{fig:training_set}, WebFace42M offers the largest cleaned training data for face recognition. Compared with the MegaFace2~\cite{MF2} dataset, the proposed WebFace42M includes 3 times more identities (2M vs. 672K), and near 10 times more images (42M vs. 4.7M). Compared with the widely used MS1M~\cite{MS1M}, our training set is 20 times (2M vs. 100K) and 4 times (42M vs. 10M) more in terms of  \# identities and \# photos. According to \cite{IMDB-Face}, there are more than 30\% and 50\% noises in MegaFace2 and MS1M, while the noise ratio of WebFace42M is lower than 10\% (similar to CASIA-WebFace~\cite{CASIA-WebFace}) based on our sampling estimation. With such a large data size, we take a significant step towards closing the data gap between academia and industry.





We further provide face attribute statistics for WebFace42M. Figure~\ref{fig:face_attributes} presents the distribution of our cleaned training data in different aspects. WebFace42M covers a large range of poses (Figure~\ref{fig:pose}), ages (Figure~\ref{fig:age}) and most major races in the world (Figure~\ref{fig:race}).

\subsection{Visualization}

Figure~\ref{fig:training_faces} illustrates four random celebrities from our WebFace data. The original WebFace260M folders downloaded/detected from Internet images are very noisy, containing various wrong detections, unrelated persons, name repetitions, etc.
The proposed CAST in the next section can automatically purify each folder to obtain cleaned faces for a certain identity. Specifically, there are large hairstyle and pose variations for celebrities \emph{Kalenna Harper} in WebFace42M. Faces of \emph{Claire Ayer} and \emph{Gregg Wallace} show different expressions and yaw angles. Finally, the proposed CAST cleaning pipeline covers a broad age range for celebrities \emph{Shaun Weiss}.
The great diversity of WebFace42M guarantees its quality for training high-performance face recognition models.

%



\section{Cleaning Automatically by Self-Training}
\label{sec:cast}

Since the images downloaded from the web are considerably noisy, it is necessary to perform a cleaning step to obtain high-quality training data. Original MS1M~\cite{MS1M} does not perform any dataset cleaning, resulting in a near 50\% noise ratio, and significantly degrades the performance of the trained models. VGGFace~\cite{VGGFace}, VGGFace2~\cite{VGGFace2} and IMDB-Face~\cite{IMDB-Face} adopt semi-automatic or manual cleaning pipelines, which require expensive labor efforts. It becomes difficult to scale up the current annotation size to even more identities. Although the purification in MegaFace2~\cite{MF2} is automatic, its procedure is complicated and there are considerably more than 30\% noises~\cite{IMDB-Face}. Another relevant exploration is to cluster faces via unsupervised approaches~\cite{otto2017clustering,kmeans} or supervised graph-based algorithms~\cite{GCND,GCNV}. However, these methods assume the whole dataset is curated, which is not suitable for the extremely noisy WebFace260M.

\begin{figure*}[t]
\centering
\includegraphics[width=0.8\linewidth]{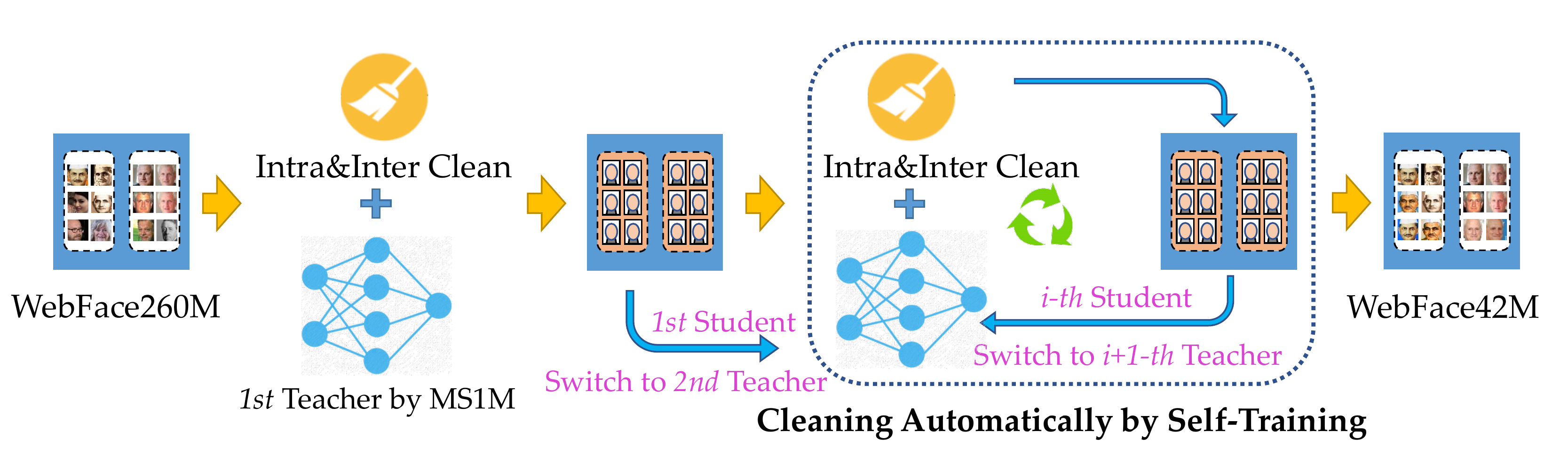}
\vspace{-2mm}
\caption{The proposed Cleaning Automatically by Self-Training (CAST). Firstly, an initial Teacher trained with MS1MV2 is utilized to clean WebFace260M. Then a Student model is trained on the cleaned WebFace data. The CAST is performed by switching the Student as the Teacher until high-quality 42M faces are obtained. Every intra-class/inter-class cleaning is conducted on the initial WebFace260M utilizing different Teacher models.}
\label{fig:self_training}
\end{figure*}

\subsection{CAST Framework}

Recently, self-training~\cite{noisy_student, yalniz2019billion}, a standard approach in semi-supervised learning, is explored to significantly boost the performance of image classification. Different from close-set ImageNet classification~\cite{ImageNet}, directly generating pseudo labels on open-set face recognition is impractical. Considering this inherent limitation, we carefully design the pipeline of Cleaning Automatically by Self-Training (CAST). Our first insight is performing self-training on open-set face recognition data, which is a scalable and efficient cleaning approach. Secondly, we find embedding feature matters in cleaning ultra-large-scale noisy face data.

\begin{figure*}
\small
\centering
\subfigure[Initial]{
\label{fig:0_score}
\includegraphics[width=0.21\textwidth]{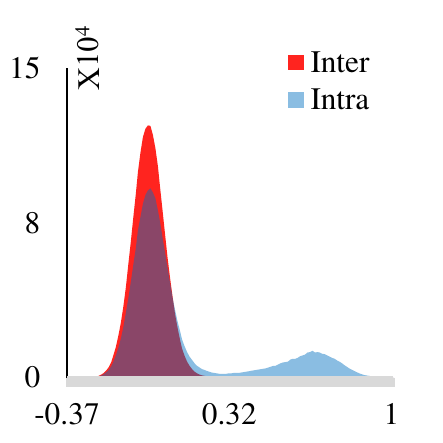}}
\subfigure[1st-iter]{
\label{fig:1_score}
\includegraphics[width=0.21\textwidth]{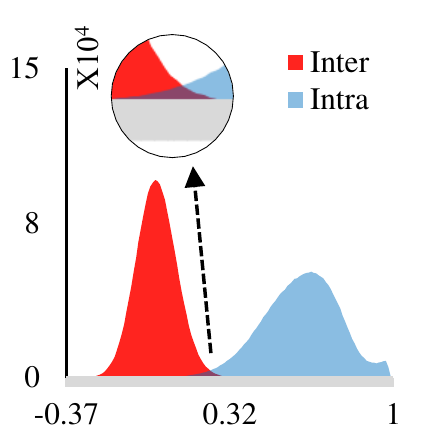}}
\subfigure[2nd-iter]{
\label{fig:2_score}
\includegraphics[width=0.21\textwidth]{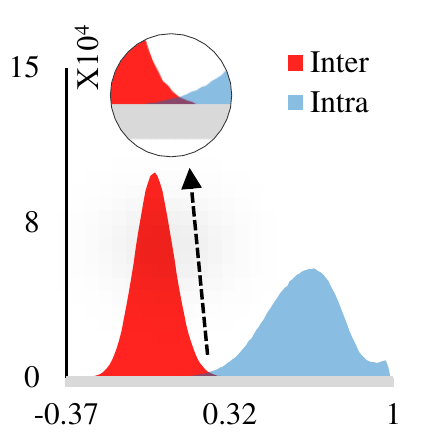}}
\subfigure[3rd-iter]{
\label{fig:3_score}
\includegraphics[width=0.21\textwidth]{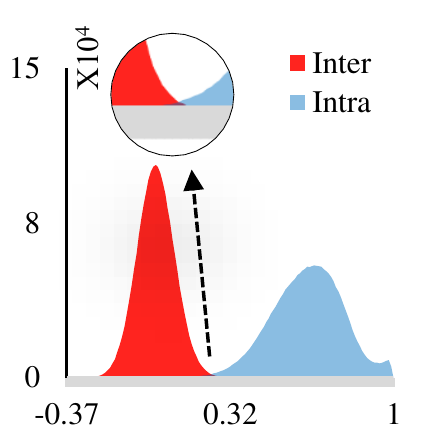}}
\caption{Inter and intra class similarity distributions during different stages of CAST. Since initial folders are very noisy, score distributions are severely overlapped. Cleaner training set is obtained after more iterations. 100K folders are randomly selected here for showing the statistic changes during iterations.}
\vspace{-5mm}
\label{fig:intrainterclean}
\end{figure*}


The overall CAST framework is shown in Figure~\ref{fig:self_training}.
Following the self-training pipeline, (1) A Teacher model is trained with the public dataset (MS1MV2~\cite{ArcFace}) to clean the original 260M images, which mainly consists of intra-class and inter-class cleaning. (2) A Student model is trained on cleaned images from (1). Since the data size is much larger, this Student generalizes better than the Teacher. (3) We iterate this process by switching the Student as the Teacher until high-quality 42M faces are obtained. It is worth noting that each intra-class/inter-class cleaning is conducted on the initial WebFace260M by different Teacher models.
All Teacher/Student models adopt ResNet-100 backbone and ArcFace loss function, and other configurations are equivalent to those in WebFace42M training setting (Section~\ref{sec:im_details}).
The CAST pipeline is summarized in Algorithm~\ref{alg:sample}.

\subsection{Intra-class and Inter-class Cleaning}
Since WebFace260M contains various noises such as outliers in a folder and identity overlaps between folders, it is impractical to perform unsupervised or supervised clustering on the whole dataset. Based on the observation that the image search results from Google are sorted by relevance and there is always a dominant subject in each search, the initial folder structure provides strong priors to guide the cleaning strategy: one folder always contains a dominant subject and different folders may contain considerable overlapped identities.

Following these priors, we perform dataset cleaning by a two-step procedure: Firstly, face clustering is parallelly conducted in 4M folders (subjects) to select each dominant identity.  Specifically, for each face in a folder, 512-dimensional embedding feature is extracted by the Teacher model, and then DBSCAN \cite{ester1996density} is utilized to cluster faces in this folder. Only the largest cluster (more than 2 faces) in each fold is reserved. $\epsilon$ and $n$ of DBSCAN indicate the maximum distance for the radius of a neighborhood, and the minimum number of points required within this distance, respectively.
We use ($1-\epsilon$) to denote the similarity of face embeddings in our paper.
With more iterations of CAST, the model learns stronger face embeddings.
A higher similarity ($1-\epsilon$) has a trend to filter out more number of noisy faces, which is beneficial for creating cleaner datasets.
So we empirically set larger values for similarity ($1-\epsilon$) in later iterations of CAST and keep $n$ fixed.
We also investigate other different designs of CAST in Section~\ref{Comparisons_of_Data_Cleaning}. Secondly, we compute the feature center of each subject to perform inter-class cleaning. Two folders are merged if their cosine similarity is higher than 0.7, and the folder containing fewer faces would be deleted when the cosine similarity is between 0.5 and 0.7.  As shown in Algorithm~\ref{alg:sample}, lines 2-5 and lines 6-13 are intra-class and inter-class cleaning processes, respectively.

\begin{algorithm}[htbp]
 \caption{Cleaning Automatically by Self-Training}
 \begin{algorithmic}[1]
 \renewcommand{\algorithmicrequire}{\textbf{\emph{1st} Model:}}
 \Require The ArcFace model trained on MS1MV2
 \renewcommand{\algorithmicrequire}{\textbf{Input Data:}}
 \renewcommand{\algorithmicensure}{\textbf{Output Data:}}
 \Require (Noisy) WebFace260M
 \Ensure  (Cleaned) WebFace42M
  \For{$i=1$ to $M$}
     \For{Each folder in WebFace260M}
     \State \parbox[t]{0.85\linewidth} {Utilizing \emph{i-th} Teacher model to extract 512-$d$ feature of each face}
     \vspace{0.5mm}
     \State \parbox[t]{0.85\linewidth} {Performing DBSCAN and reserving the largest cluster to obtain intra-class cleaning results}
     \EndFor
     \For{Every two subjects}
     \State Computing the cosine similarity of feature center
     \If {cosine similarity \textgreater 0.7}
        \State Merging two folders
     \ElsIf {cosine similarity \textgreater 0.5}
        \State Deleting the folder with fewer faces
     \EndIf
     \EndFor


     \State \parbox[t]{0.88\linewidth} {Training \emph{i-th} Student model on \emph{i-th} cleaned WebFace (not for last iteration)}
     \vspace{0.5mm}
     \State \parbox[t]{0.88\linewidth} {Converting \emph{i-th} Student into \emph{i+1-th} Teacher model (not for last iteration)}
  \EndFor

\State \Return Cleaned WebFace data
\end{algorithmic}
\label{alg:sample}
\end{algorithm}

The effectiveness of the above intra-class and inter-class cleaning heavily depends on the quality of the embedding feature, which is guaranteed by the proposed self-training pipeline. The ArcFace model trained on MS1MV2 with ResNet-100 provides a good initial embedding feature to perform first-round cleaning for WebFace260M. Then, this feature is significantly enhanced with more training data in later iterations. Figure~\ref{fig:intrainterclean} illustrates the score distribution during different stages of CAST, which indicates a cleaner training set after more iterations. Furthermore, the ablation study in Table~\ref{tab:castcleaningresult} also validates the effectiveness of the CAST pipeline. It is worth noting that the proposed CAST pipeline is compatible with any intra-class and inter-class strategies.


\noindent{\bf Remove duplicates and test set overlaps.}
After CAST, duplicated faces are removed when their cosine similarity is higher than 0.95. Furthermore, the feature center of each subject is compared with popular benchmarks (such as LFW families~\cite{LFW,CALFW,CPLFW}, FaceScrub~\cite{FaceScrub}, IJB-C~\cite{IJB-C}, the proposed test set, etc.), and overlaps are removed if the cosine similarity is higher than 0.7.

\begin{figure}
\small
\centering
\includegraphics[width=0.47\textwidth]{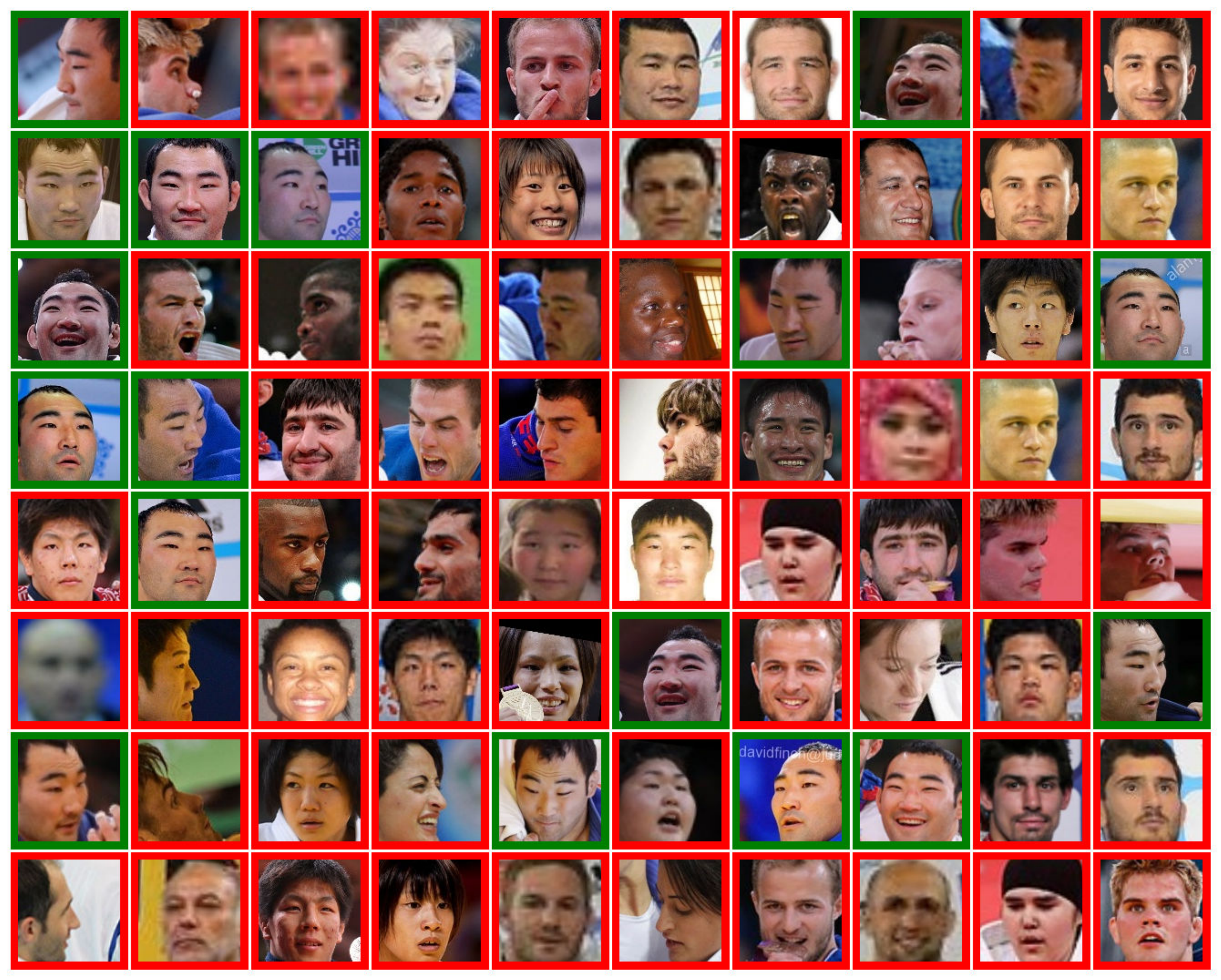}
\vspace{-2mm}
\caption{Illustration of the reserved and the rejected face samples. \textcolor{green}{Green} and \textcolor{red}{red} boxes denote reserved faces and faces rejected by our CAST, respectively.}
\vspace{-5mm}
\label{fig:rejection_2}
\end{figure}

Figure~\ref{fig:rejection_2} illustrates the reserved and the rejected face samples in our cleaning process. One can find that noisy and mislabeled faces are successfully rejected by the proposed CAST strategy, and most true positives are reserved in the cleaned WebFace42M set. Significantly, there are diverse expressions and poses among the remaining faces, which clearly show the effectiveness of our CAST and the high quality of our training data.


\begin{table}[t]
\caption{The identities and images statistics during different cleaning stages.}
\begin{center}{\scalebox{1.0}{
\begin{tabular}{l|c|c|c}
\hline
\multicolumn{2}{c|}{Stages} & \# Identities & \# Faces \\ \hline
\hline
\multicolumn{2}{c|}{Collect name list and images} & 4,073,509 & 265,777,598   \\
\multicolumn{2}{c|}{Face pre-processing} & \textbf{4,008,130} & \textbf{260,890,076} \\ \hline
\multirow{2}{*}{First iteration}
& Intra-class & 3,341,761 & 61,792,387 \\
& Inter-class & 2,437,140 & 50,672,354 \\ \hline
\multirow{2}{*}{Second iteration}
& Intra-class & 3,027,814 & 60,274,892 \\
& Inter-class & 2,176,427 &47,352,741 \\ \hline
\multirow{2}{*}{Third iteration}
& Intra-class & 2,878,886 & 58,155,345 \\
& Inter-class &2,070,870  & 46,220,417 \\ \hline
\multicolumn{2}{c|}{Remove duplicates} & 2,070,870 &43,977,802 \\
\multicolumn{2}{c|}{Remove test set overlaps} & \textbf{2,059,906} & \textbf{42,474,558}\\ \hline
\end{tabular}}}
\end{center}

\label{table:cast_sta}
\end{table}

\begin{figure*}[t]
\centering
\includegraphics[width=0.92\linewidth]{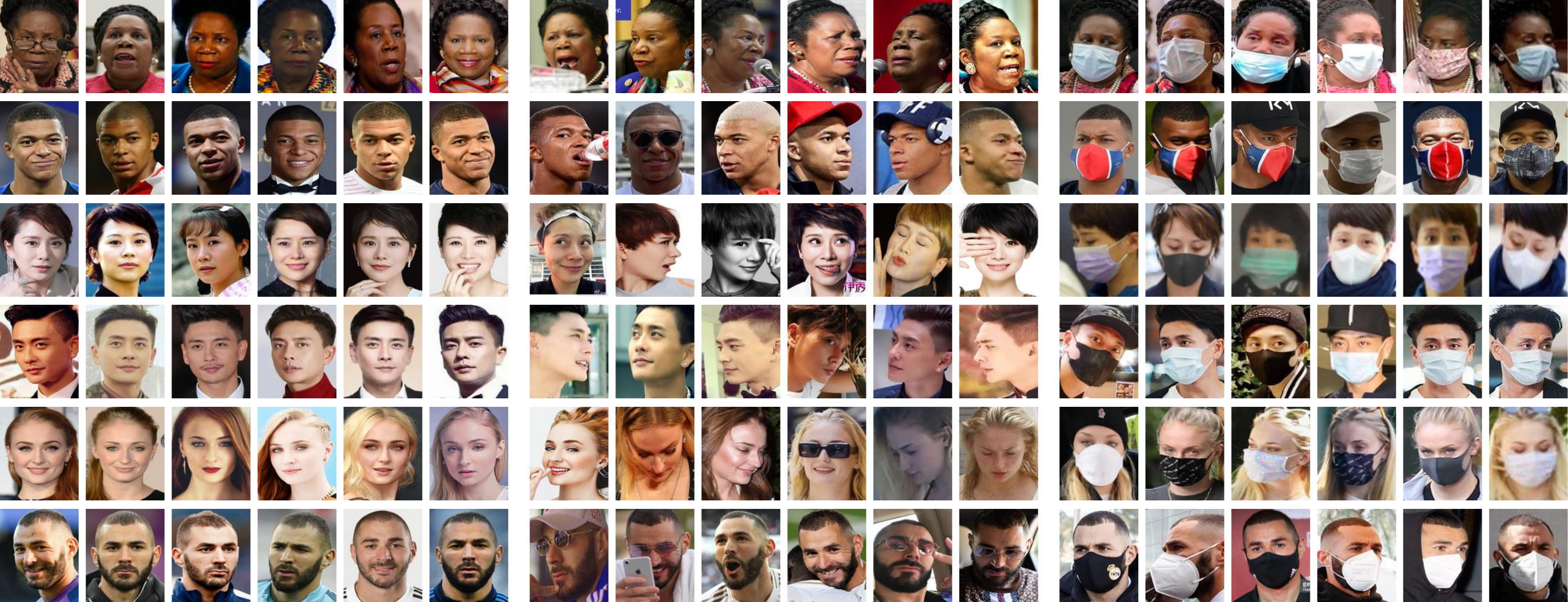}
\caption{Visualization of our test set. Random faces of a certain identity
are shown in each row. 1-6 columns: Controlled, 7-12 columns: Wild, 13-18 columns: Masked.
}

\label{fig:mask_face}
\end{figure*}

\subsection{Statistics}
The statistics of identities and faces during different cleaning stages are shown in Table~\ref{table:cast_sta}. After face pre-processing for downloaded images, there are 4,008,130 identities and 260,890,076 faces (WebFace260M). The face set becomes cleaner under more CAST iterations, which results in fewer data. Finally, we obtain 2,059,906 identities and 42,474,558 faces (WebFace42M) after removing duplicates and test set overlaps.

\section{Face Recognition Evaluation}

In this section, we firstly introduce the time-constrained face recognition evaluation protocol, which covers various practical applications. Then, three benchmarking tasks (standard, masked, unbiased face recognition) are detailed, including corresponding background, test sets, and metrics.

\subsection{FRUITS Protocol}

As discussed in Section~\ref{recognition_evaluation}, most existing face recognition evaluations \cite{LFW,CFP,YTF,MegaFace,IJB-C} only focus on accuracy.  LFR Challenge \cite{LFR} adopts FLOPs and model size constraints, which may result in unfair inference time comparisons. Moreover, it ignores the computation module of face detection and alignment.  Four months submission interval of NIST-FRVT \cite{FRVT} makes it difficult for the research community to freely perform experiments.
In this paper, we design the Face Recognition Under Inference Time conStraint (FRUITS) protocol, which enables academia to comprehensively evaluate their face matchers. Referring to~\cite{FRVT},  inference time is measured on a single core of an Intel Xeon CPU E5-2630-v4@2.20GHz processor. For face recognition evaluation with flip, the batch size is set as 2.
Considering different application scenarios, the FRUITS protocol sets a series of tracks:

\noindent{\bf {FRUITS-100}}: The whole face recognition system must distinguish image pairs within 100 milliseconds, including pre-processing (face detection and alignment), feature embedding for recognition, and matching. FRUITS-100 track targets on evaluating lightweight face recognition systems which can be deployed on mobile devices.

\noindent{\bf{FRUITS-500}}: This track follows FRUITS-100 setting, except that the time constraint is increased to 500 milliseconds. This track aims to evaluate modern and popular networks deployed in the local surveillance system.

\noindent{\bf{FRUITS-1000}}: Following NIST-FRVT, FRUITS-1000 constrains the inference time in 1000 milliseconds and aims to compare capable recognition models performed on clouds.
In our organized Face Bio-metrics under COVID Workshop, FRUITS-1000 is adopted for both masked and standard face recognition challenges.

\begin{table*}[t]
\caption{The statistics of our test set, including SFR, MFR, and UFR evaluations. - means corresponding statistics or comparisons are omitted.}
\begin{center}{\scalebox{0.92}{
\begin{tabular}{c|c|c|c|c|c|c}
\hline
Evaluation &
\multicolumn{2}{c|}{Attributes}& \# Identities & \# Faces & \# Impostor & \# Genuine \\ \hline
\hline

\multirow{6}{*}{\tabincell{l}{Standard Face Recognition (SFR)}}

&\multicolumn{2}{c|}{\textbf{All}} & \textbf{2,478} & \textbf{57,715} & \textbf{1,664,475,460} & \textbf{1,006,295}\\
\cline{2-7}

&
\multirow{2}{*}{\tabincell{l}{Age}}

&Cross-age-10& - & - & 833,518,127 & 276,587\\
&&Cross-age-20& - & - & 8,421,183,318 & 61,780 \\ \cline{2-7}

&
\multirow{3}{*}{\tabincell{l}{Scenarios}}

& Controlled& - & 22,135 & 244,828,795  & 139,250\\
&&Wild& - &  35,580   & 632,449,086 & 501,324\\
&&Cross-scene& - & - & 787,197,579 & 365,721   \\  \hline

\multirow{6}{*}{\tabincell{l}{Masked Face Recognition (MFR)}}

&
\multicolumn{2}{c|}{\textbf{All}} & \textbf{2,478} & \textbf{60,926} & - & -\\
\cline{2-7}

&

\multicolumn{2}{c|}{Masked} & 862  & 3,211 & - & -\\
\cline{2-7}

&

\multicolumn{2}{c|}{Nonmasked} & 2,478 & 57,715 & - & -\\
\cline{2-7}

%

&
\multirow{3}{*}{\tabincell{l}{Mask}}

& Controlled-Masked& - & - & 71,042,982  & 32,503  \\
&&Wild-Masked& - &  -   & 114,193,476 & 53,904\\
&&All-Masked& - & - & 185,236,458 & 86,407   \\  \hline

%





\multirow{6}{*}{\tabincell{l}{Unbiased Face Recognition (UFR)}}
&
\multirow{4}{*}{\tabincell{l}{Race}}

&Caucasian& 990 & 22,940  & 262,676,189  & 434,141 \\
&&East Asian& 874 & 22,970   & 263,392,669 & 406,296 \\
&&African& 447 & 8,528 & 36,234,242 & 124,886 \\
&&Others&167 & 3,277 & - & - \\ \cline{2-7}

&
\multirow{2}{*}{\tabincell{l}{Gender}}

&Male& 1527 & 33,872 & 573,108,363 & 530,893  \\
&&Female& 951 & 23,843 & 283,757,001 & 475,402 \\ \hline

\end{tabular}}}
\end{center}

\vspace{-4mm}
\label{table:test_set_num}
\end{table*}


\subsection{Standard Face Recognition}
\label{sec:test_set}

\subsubsection{Test set}
Since public evaluations are most saturated and may contain noises,
we manually construct an elaborated test set for SFR, MFR and UFR. It is well known that recognizing strangers, especially when they are similar-looking, is a difficult task even for experienced vision researchers. Therefore, our multi-ethnic annotators only select their familiar celebrities, which ensures the high quality of the test data.
Besides, annotators are encouraged to gather attribute-balanced faces, and \textbf{recognition models are introduced to guide hard sample collection}.
 The statistics of the final test set are listed in Table~\ref{table:test_set_num}. In total, there are 60,926 faces of 2,478 identities. Rich attributes (such as age, race, gender, scenario) are accurately annotated. Among all collected data, 57,715 faces are utilized for SFR evaluation.
 In the future, we will actively maintain and update this test set.
In Figure~\ref{fig:mask_face}, we show the samples of our SFR test set, including \emph{Controlled} and \emph{Wild} faces. One can find that there is great diversity in ages, poses, scenarios, etc.

\subsubsection{Metrics}
Based on the proposed FRUITS protocol and test set, we perform standard 1:1 face verification across various attributes. Table~\ref{table:test_set_num} shows numbers of imposter and genuine in different verification settings. \emph{All} means impostors are paired without attention to any attribute, while later comparisons are conducted on age and scenario sub-sets. \emph{Cross-age} refers to cross-age (more than 10 or 20 years span) verification, while \emph{Cross-scene} means pairs are compared between \emph{Controlled} and \emph{Wild} settings. Different algorithms are measured on False Non-Match Rate (FNMR) \cite{FRVT}, which is defined as the proportion of mated comparisons below a threshold set to achieve the False Match Rate (FMR) specified. FMR is the proportion of impostor comparisons at or above that threshold. \textbf{Lower FNMR at the same FMR is better}.

\subsection{Masked Face Recognition}

\subsubsection{Face Bio-metrics under COVID}
SFR systems usually work with mostly non-occluded faces, which include primary facial features such as eyes, nose, and mouth. However, there are a number of circumstances in which faces are occluded by masks such as in pandemics, medical settings, excessive pollution, or laboratories. According to WHO statistics, there are more than 235,408,082 confirmed COVID-19 cases including 4,809,149 deaths worldwide till October 6, 2021.
During the coronavirus epidemic, almost everyone wears a facial mask, which poses a huge challenge to face recognition. Traditional SFR may not effectively recognize the masked faces, but removing the mask for authentication would increase the virus infection risk.

To cope with the above-mentioned challenging scenarios arising from wearing masks, it is crucial to improve the existing face recognition approaches. Recently, some commercial vendors \cite{FRVT-mask} have developed face recognition algorithms capable of handling face masks, and an increasing number of research publications \cite{ding2020masked,geng2020masked,du2021towards,hariri2021efficient,anwar2020masked} have surfaced on this topic. However, due to the sudden outbreak of the epidemic, there is yet no publicly available large-scale MFR benchmark.

\subsubsection{Test Set with Real-world Mask}
In contrast with simulated \cite{FRVT-mask, negi2021deep} or relatively small \cite{anwar2020masked,damer2021extended,IJCB-mask,wang2020masked} masked test sets, a real-world comprehensive benchmark for evaluating MFR is developed in this work. Based on the SFR identities, we further collect masked faces for these celebrities. Specifically, as shown in Table \ref{table:test_set_num}, there are carefully selected 3,211 masked faces among 862 identities. Subjects with real-world masks are illustrated in Figure~\ref{fig:mask_face}. Wearing masks causes severe occlusion, resulting in just the periocular area and above visible. Besides, there are changeful mask types, colors, wearing ways, and head poses in real-world applications, which are more practical and challenging than simulated ones.

For MFR, assessment is performed with \emph{Mask-Nonmask} comparisons. Specifically, there are one masked face and another face from standard face sets for pair verification. According to the attributes of faces without masks, we evaluate the performance of algorithms under \emph{Controlled-Masked}, \emph{Wild-Masked}, and \emph{All-Masked} settings in Table \ref{table:test_set_num}.



%
%


\subsection{Unbiased Face Recognition}

\subsubsection{Bias and Fairness in Face Biometrics}
Bias in face recognition means it provides higher accuracy within certain demographic groups and lower performance for other demographics. According to the NIST-FRVT report \cite{grother2019face}, most submitted recognition algorithms from academia and industry exhibit different levels of biased
performances. Deploying such systems may cause significant consequences such as racism.
Recent UFR researches \cite{RFW,gong2020jointly,wang2019mitigate,terhorst2021comprehensive} mainly focus on balanced data collection/sampling and de-biased algorithm design. For evaluation, most of the tests are performed on RFW set \cite{RFW}, which adopts LFW-like pair comparisons. In this paper, based on the proposed test data and FRUITS protocol, we provide a more challenging and practical UFR evaluation.

\subsubsection{Test Set and Metrics}
To enable a trustworthy face recognition system, it is of importance to investigate the performance on different facial attributes.
As shown in Table~\ref{table:test_set_num}, the test set of UFR is the same as SFR ones.
We manually label race and gender attributes for unbiased evaluation.  Fairness assessment of ethnicity is reported on Caucasian, East Asian, African, and so does gender. Following common practices \cite{wang2019mitigate} in the community, we adopt skewed error ratio (SER) and standard deviation (STD) as the fairness metrics.
Specifically, the error ratio of each race and gender attribute is calculated according to FNMR@FMR=1e-5. Then, SER can be computed by the ratio of the highest to the lowest error rate among race and gender groups.  STD is the number of error dispersion among different races and genders.

\section{Million-level Face Recognition Experiments}

Based on the constructed WebFace260M benchmark, we dive deep into the million-scale face recognition in this section. Firstly, implementations are detailed containing parameter and environment configurations. Then we analyze the speed and performance of distributed framework, which enables large-scale face recognition training. Thirdly, the WebFace data is compared with public counterparts, covering different losses and test sets. Furthermore, the proposed CAST strategy and its key procedures are studied. Lastly, we establish comprehensive baselines for SFR, MFR, UFR, and report the performance on NIST-FRVT.

\subsection{Implementation Details}
\label{sec:im_details}

\noindent{\bf{Hyper-parameters.}} In order to fairly evaluate the performance of different face recognition models, we reproduce representative algorithms (CosFace \cite{CosFace}, ArcFace \cite{ArcFace} and CurricularFace \cite{Curricularface}) in one Gluon codebase.
Margin values in CosFace~\cite{CosFace}, ArcFace~\cite{ArcFace} and CurricularFace~\cite{Curricularface} are set as 0.35, 0.5 and 0.5, respectively. The Stochastic Gradient Descent (SGD) with momentum 0.9 and weight decay 0.0005 is utilized for the network optimization. For large-batch training with cluster, we employ distributed synchronous SGD, which parallelizes the tasks across machines.
The default batch size per GPU is set as 64 unless otherwise indicated. The learning rate is set as 0.05 for a single node (8 GPUs), and follows the linear scaling rule \cite{goyal2017accurate} for the training on multiple nodes ( $0.05\times$\# machines). We decrease the learning rate by 0.1$\times$ at 8, 12, and 16 epochs, and stop at 20 epochs for all models.
Gradual warmup~\cite{goyal2017accurate} is adopted during the initial 1 epoch and 5 epochs for a single node and multiple nodes training, respectively.
During training, we only adopt horizontal flip data augmentation.
In DBSCAN, similarity ($1-\epsilon$) is set as 0.5, 0.55, and 0.6 for 1st, 2nd, and 3rd iterations, respectively. $n$ is set as 3 for all iterations.

\begin{table}
\caption{Speed and performance comparisons of distributed training.   ResNet-100 backbone with ArcFace loss is adopted. B, G and M refer to batch size per GPU, \# GPUs per machine, and \# machines. $X$ and $W$ mean feature and center, and numbers in bracket are the GPU memory usage (MB). Performance is reported on IJB-C (TAR@FAR=1e-4).}
\begin{center}{\scalebox{0.9}{
\begin{tabular}{l|c|c|c|c|c|c}
\hline
 Data & B$\times$G$\times$M &FP32/16 &Parallel & Speed  & Time & IJB-C \\
\hline
\hline
\multirow{6}{*}{\tabincell{c}{10\%}}
& 32$\times$8$\times$1 & FP32 & $X$ (7913)&0.6K& 39h & 96.67 \\
& 64$\times$8$\times$1 & FP32 & $X$ $W$ (7521)  &0.9K&26h& 96.83 \\
& 64$\times$8$\times$1 & FP16 & $X$ (7551) &1K& 23h & 96.80 \\
& 64$\times$8$\times$1 & FP16 & $X$ $W$ (7182)&1.8K& 13h & 96.78 \\
& 64$\times$8$\times$4 & FP16 & $X$ $W$ (7125) &6.3K& 4h  & 96.73 \\
& 64$\times$8$\times$8 & FP16 & $X$ $W$ (7119) &12.4K& 2h  & 96.77 \\
\hline
\multirow{3}{*}{\tabincell{c}{30\%}}
& 64$\times$8$\times$1 & FP16 & $X$ $W$ (8901)&1.7K&41h & 97.41\\
& 64$\times$8$\times$4 & FP16 & $X$ $W$ (8519)&5.5K&13h &97.50\\
& 64$\times$8$\times$8 & FP16 & $X$ $W$ (8455)&11.3K&6h &97.47 \\
\hline
\multirow{4}{*}{\tabincell{c}{100\%}}
& 32$\times$8$\times$1 & FP16 & $X$ $W$ (10503)&1K& 233h & 97.71 \\
& 32$\times$8$\times$8 & FP16 & $X$ $W$ (8359)&6.8K&34h & 97.65\\
& 32$\times$8$\times$16 & FP16 & $X$ $W$ (8297)&12.9K&18h & 97.74\\
& 32$\times$8$\times$32 & FP16 & $X$ $W$ (8221)&25.3K& \textbf{9h}& 97.70 \\
\hline
\end{tabular}}}
\end{center}

\label{table:distributed_training}
\end{table}

\begin{figure}[t]
\centering
\includegraphics[width=1.0\linewidth]{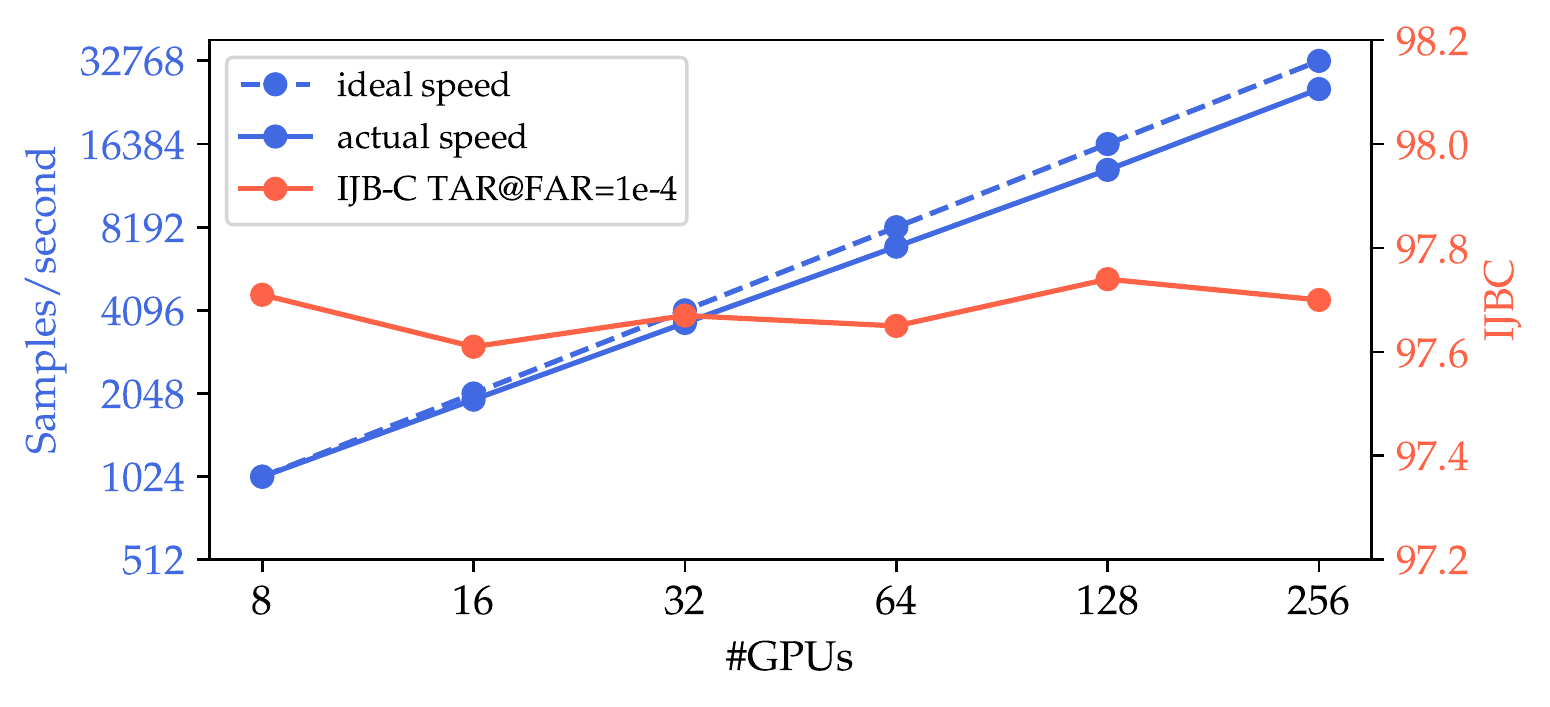}
\vspace{-0.4cm}
\caption{Speed and performance of our distributed training system, which achieves almost linear acceleration with comparable performance. 100\% data (WebFace42M) is used in this experiment.}
\label{fig:distributed_training}
\vspace{-0.2cm}
\end{figure}

\noindent{\bf{Hardware and software.}} All experiments are performed on a cluster containing 35 nodes, and each node contains 8 GPUs, 2 CPU processors, and 384GB RAM memory. Machines are inter-connected with a 30 Gbps TCP/IP network. For storage, the cluster has 32TB distributed SSDs. For face recognition training, CUDA version 10.0 with cuDNN version 7.6.0 is employed. We use NCCL version 2.4.7 as the communication library.



\subsection{Distributed Training}

When using the ultra-large-scale WebFace42M as the training data and computationally demanding backbones as the embedding networks, the model optimization can take several weeks on a single machine. Such a long training time makes it difficult to efficiently perform experiments. Inspired by the distributed optimization on ImageNet~\cite{goyal2017accurate}, we apportion the workload of model training to clusters. To this end, parallelization on both feature $X$ and center $W$,  mixed-precision (FP16) and large-batch training are adopted in this work.

\begin{table*}[t]
\caption{Performance comparisons of our WebFace and public training data. ResNet-100 backbone \textbf{without flip test} is adopted. RFW refers to average accuracy on~\cite{RFW}, MegaFace refers to rank-1 identification and verification scores on \cite{MegaFace}, IJB-C is TAR@FAR=1e-5 and 1e-4 on~\cite{IJB-C}. On our test set, FNMR@FMR=1e-5 on \emph{All} pairs of SFR, FNMR@FMR=1e-5 on \emph{All-Masked} pairs of MFR, race SER metric of UFR are reported (\textbf{lower is better}). The best results are marked in bold.}
\centering
\begin{center}{\scalebox{0.88}{
\begin{tabular}{c|c|c|c|c|c|c|c|c|c|c|c|c|c|c}
\hline
\multirow{2}{*}{Data} & \multirow{2}{*}{Loss} & \multirow{2}{*}{LFW} &\multirow{2}{*}{CALFW} &\multirow{2}{*}{CPLFW} &\multirow{2}{*}{AgeDB30} & \multirow{2}{*}{CFP-FP} &\multirow{2}{*}{RFW} & \multicolumn{2}{c|}{MegaFace} &\multicolumn{2}{c|}{IJB-C} & \multicolumn{3}{c}{Our test set $\downarrow$} \\

\cline{9-15}

&&&&&&&& Id.  & Veri. & 10-5  & 10-4 & SFR & MFR & UFR \\



\hline \hline

\multirow{3}{*}{MS1M}
& CosFace    &98.32&95.19&91.84&96.01& 97.09 &98.09& 96.21 &97.73 &86.48& 92.96  &62.91 & 89.36 & 2.37 \\
& ArcFace   &98.22 &94.05&91.59&96.73&97.07 &97.64& 97.67 &97.94&89.03& 93.45 &  61.73 &86.42 & 1.84\\
& CurricularFace   & 98.89 &95.46&91.18&96.82&96.21  &98.12 & 96.86 &97.26&85.47& 92.99 & 64.05 &87.78&1.65\\
\hline

\multirow{3}{*}{MS1M-IBUG}
& CosFace    &99.76 &95.83&89.98&94.95&97.83 &97.62& 97.33 &97.60&89.79& 94.35 &  12.22&71.41& 1.32\\
& ArcFace    &99.76&95.91&89.41&94.40&97.95 &97.78& 97.27 &97.35& 91.50 &  94.57  & 13.46&74.61&1.49\\
& CurricularFace   &99.78  &95.95&90.37&94.51&97.94& 97.86& 97.19 &97.24&89.85& 94.72 & 13.35&75.05& 1.43\\
\hline
\multirow{3}{*}{MS1MV2}
& CosFace    &99.81 &96.18&92.76&98.34&98.18 &98.85& 98.30& 98.33&94.62&96.01 &  9.04&68.25&1.34 \\
& ArcFace   &99.78&96.05&92.93&98.21& 98.54&98.98&98.40& 98.24 &94.05& 96.03 & 9.88&71.81&1.40\\
& CurricularFace   &99.83 &96.28&93.05&98.32&98.67  & 99.02&98.46 & 98.47&94.01& 96.21 & 9.52&69.56&1.38\\
\hline
\multirow{3}{*}{MS1M-Glint}
& CosFace    &99.83 &96.00&91.03&94.95&98.16 &99.59&98.60&98.83 &94.00&96.15 &  12.65 &74.33&2.00\\
& ArcFace    &99.71 &95.96&90.76&94.20&98.43 &99.60&98.48&98.31 &93.35&96.24 &  13.98 &78.09&2.19 \\
& CurricularFace   &99.80 &\textbf{96.31}&91.65&95.47 &98.82& \textbf{99.65}&98.57&98.60 &93.20&96.31  & 14.27 &74.27&2.12 \\
\hline

\multirow{3}{*}{MegaFace2}
& CosFace   &99.46 &93.25&92.46&87.67& 89.76&88.90&86.62&89.13 &79.75&87.75 & 58.77 &92.42& 1.55\\
& ArcFace   &99.50 &92.86&88.60&91.06&93.88 &89.45&88.28&90.68 &81.75& 89.35  & 54.73 &91.54& 1.57\\
& CurricularFace   &99.48  &93.02&90.39&91.07&93.04 &90.06 & 88.32 &90.00 &82.27& 90.11  & 55.73 &91.65&1.56 \\
\hline

\multirow{3}{*}{IMDB-Face}
& CosFace     &99.61 &95.08&92.15&96.75& 98.48&93.80&94.03&95.12 &89.92&93.96  &27.88 &87.56&2.92\\
& ArcFace   &99.58 &94.85&91.83&97.29& 98.45&93.08&93.48&94.62 &89.09& 93.37 & 30.45 &90.25&3.17\\
& CurricularFace   &99.73 &94.81&92.44&97.46&98.66 &94.11&93.63 &94.68 &88.78& 94.12 & 30.94&88.54&2.95\\
\hline

\multirow{3}{*}{WebFace4M}
& CosFace   &99.80 &95.95&94.40&97.45&99.25 &98.16&97.59&98.32 &94.93&96.86  &  8.98&68.77&1.49 \\
& ArcFace   &\textbf{99.85} &95.93&94.31&97.82&99.04 &98.14&97.60&98.00&94.72&96.77  &  10.39 &70.97&1.45\\
& CurricularFace   &99.83 &96.03&94.21&97.83& 99.11&98.14&97.94&98.19 &95.15&97.02  & 9.09 &66.58&1.51\\
\hline

\multirow{3}{*}{WebFace12M}
& CosFace   &99.81  &96.18&94.80&97.88&\textbf{99.38}&99.15&98.66 &98.75&96.00&97.41  & 4.34 &54.08&1.41\\
& ArcFace   &99.81 &96.11&94.68&98.33& 99.37&99.08&98.82 &98.63 &95.96&97.47 & 4.77 &56.47&1.39\\
& CurricularFace   &99.83 &96.13&94.70&98.37&99.37 &99.18&98.75 &\textbf{98.85}&95.99&97.51  & 4.81 &53.05&1.46\\

\hline

\multirow{3}{*}{WebFace42M}

& CosFace  & 99.83 &96.11&94.90&\textbf{98.58}& \textbf{99.38}& 99.41&99.02&98.57 &\textbf{96.44}&97.68  & 3.01 &45.28&1.31 \\
& ArcFace  &99.83 &96.19&\textbf{94.93}&98.02& 99.28&99.33&99.02& 98.61&96.23&97.70 & \textbf{2.98} &47.25&\textbf{1.30}   \\
& CurricularFace   &99.83  &96.10&94.85&98.24& \textbf{99.38}& 99.39& \textbf{99.11} &98.55&96.38&  \textbf{97.76} &3.05 &\textbf{42.97}&1.33\\

\hline
\end{tabular}}}
\end{center}

\label{tab:trainingdatavs}
\end{table*}

\begin{table}[t]

\caption{Performance (\%) of ArcFace models trained with ResNet-14 on different portions of WebFace data. TAR@FAR=1e-4 on IJB-C is reported.}
\centering

\begin{center}{\scalebox{1.0}{
\begin{tabular}{c|c|c|c}
\hline
 Training data &WebFace4M & WebFace12M& WebFace42M  \\ \hline\hline
 IJB-C &93.13 &93.92 & 94.22    \\
\hline
\end{tabular}}}
\end{center}

\label{table:small_model}
\end{table}

The speed and performance of our distributed training system are illustrated in Table~\ref{table:distributed_training} and Figure~\ref{fig:distributed_training}. Parallelization on both feature $X$ and center $W$ as well as mixed-precision (FP16) significantly reduce the consumption of GPU memory and speed up the training process, while similar performance can be achieved. Equipped with 8 nodes (64 GPUs), the training speed is scaled to 12K samples/s and 11K samples/s on WebFace4M (10\% data) and WebFace12M (30\% data), respectively. The corresponding training time is only 2 hours and 6 hours. Furthermore, the scaling efficiency of our training system is above 80\% when applied to ultra-large-scale WebFace42M on 32 nodes (256 GPUs).
Figure~\ref{fig:distributed_training} shows the ideal speed, actual speed, and corresponding performance for increasing GPUs resource.
We can reduce the training time of the ResNet-100 model from 233 hours (1 node) to 9 hours (32 nodes) with comparable performance.

\subsection{Comparisons of Training Data}

For comprehensively benchmarking the influence of training data, the proposed WebFace42M is compared with public counterparts including MS1M families \cite{MS1M,deng2017marginal,ArcFace,glintweb},  MegaFace2 \cite{MF2} and IMDB-Face \cite{IMDB-Face}. 10\% (WebFace4M) and 30\% (WebFace12M) random selection of our full data are also employed for further analyses. The statistics of different training sets are illustrated in Table~\ref{table:training_set}. Evaluation sets used in this experiment include popular verification sets (LFW \cite{LFW}, CALFW \cite{CALFW}, CPLFW \cite{CPLFW}, AgeDB \cite{AgeDB}, CFP-FP \cite{CFP}), RFW \cite{RFW}, MegaFace \cite{MF2}, IJB-C \cite{IJB-C} and our test set.


\begin{figure}
\centering
\includegraphics[width=1.0\linewidth]{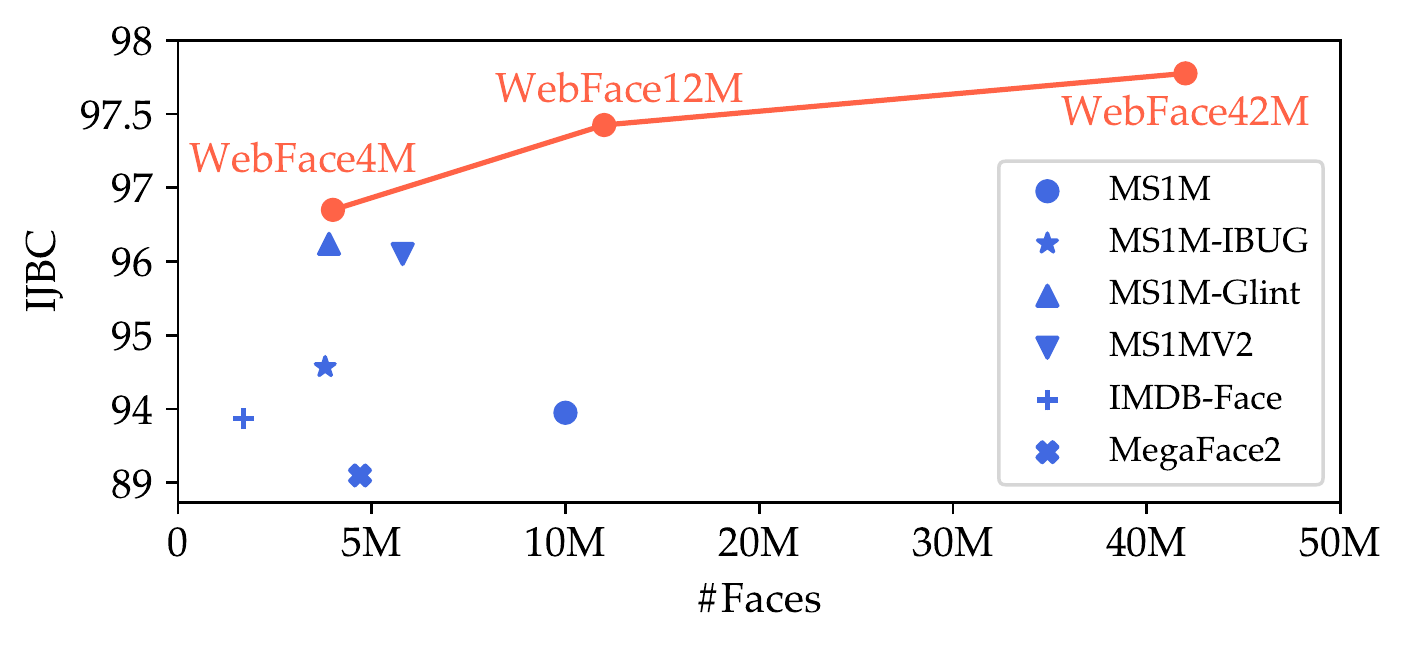}
\caption{Performance of ArcFace models (ResNet-100) trained on the WebFace envelopes counterparts trained on the public training data.}
\label{fig:training_setvs}
\end{figure}

As we can see from Table~\ref{tab:trainingdatavs} and Figure~\ref{fig:training_setvs}, the proposed WebFace42M breaks the bottleneck of training data for deep face recognition across various loss functions and test sets.
In summary, almost all best accuracies on each test set are achieved by the proposed WebFace data shown in Table~\ref{tab:trainingdatavs}.
Specifically, WebFace42M reduces 40\% error rate on the challenging IJB-C dataset compared with MS1MV2, boosting TAR@FAR=1e-4 from 96.03\% to 97.70\% with ResNet-100 and ArcFace. Along with the increment of data scale (10\%, 30\%, and 100\%), there exists a consistent improvement in performance as observed in Figure~\ref{fig:training_setvs}, enveloping its public counterparts. On MegaFace identification task, WebFace42M sets a new state of the art by 99.11\% rank-1 score.
On our test set, for \emph{All} pairs comparison of SFR, WebFace42M decreases the FNMR@FMR=1e-5 from 9.88\% to 2.98\%, reducing the error rate by more than 3 times. For \emph{All-Masked} pairs comparison of MFR, FNMR@FMR=1e-5 metric is boosted from 69.56\% to 42.97\% by the proposed training data. Meanwhile, due to the tremendous scale and diversity of WebFace, it benefits UFR according to SER metric.
Furthermore, the models trained on 10\% data, WebFace4M, impressively achieve superior performance compared to ones trained on  MS1M families and MegaFace2, which include even more faces.
Undisputedly, this comparison confirms the effectiveness and necessity of our WebFace42M in leveling the playing field for million-scale face recognition.



%

Besides reporting the face recognition results of ResNet-100, we also train ArcFace models by using a ResNet-14 network on different portions of our data (10\%, 30\% and 100\%). As given in Table~\ref{table:small_model}, there is also a consistent performance gain for ResNet-14 when more training data are progressively employed. Therefore, the proposed WebFace42M is not only beneficial to the large model (ResNet-100) but also valuable for the lightweight model. This is of significance for mobile devices like cellphones, and we explore smaller models in Section ~\ref{sec:FRUITS-100}.

\subsection{Comparisons of Data Cleaning}
\label{Comparisons_of_Data_Cleaning}

As shown in Table~\ref{tab:castcleaningresult}, the CAST pipeline is compared with other cleaning strategies on the original MS1M \cite{MS1M} and WebFace260M. Specifically, for MS1M results, the initial teacher model is trained on IMDB-Face \cite{IMDB-Face} by using ResNet-100 and ArcFace. Then, CAST is conducted on the noisy MS1M following Section~\ref{sec:cast}. After steps of iteration, our fully automatic cleaning strategy provides highly-curated data for model training, outperforming semi-automatic methods used in ~\cite{deng2017marginal,ArcFace,glintweb}. Compared with the most recent GCN-based cleaning \cite{MillionCelebs}, the data cleaned by the CAST also achieves higher performance.


\begin{table}[t]

\caption{Comparisons of CAST and other data cleaning pipelines. Pairs refers to average accuracy on~\cite{LFW,CFP,AgeDB,CALFW,CPLFW}. MegaFace and IJB-C refer to rank-1
identification and TAR@FAR=1e-4 respectively.
For MS1M and WebFace by CAST, different iterations are compared. CAST-1 means the first-round iteration.}
\centering
\begin{center}{\scalebox{0.95}{
\begin{tabular}{c|c|c|c|c|c}
\hline
Data & \# Id & \# Face & Pairs & MegaFace & IJB-C \\\hline\hline
 MS1M &100K &10M & 95.53   &97.67 & 93.45\\\hline
 MS1M-IBUG& 85K &3.8M & 95.49   &97.27 & 94.57\\
 MS1MV2& 85K &5.8M & 97.10   &98.40 &96.03\\
 MS1M-Glint& 87K &3.9M & 95.81   & 98.48&96.24 \\
 MS1M-GCN \cite{MillionCelebs} & - & - & 96.51 & - & -  \\ \hline
 MS1M by CAST-1&94K &6.3M & 95.37 & 97.93  &94.31\\
 MS1M by CAST-2&92K &5.5M & 97.08 & 98.47  &95.90\\
 MS1M by CAST-3& 91K& 4.9M& 97.42 & 98.61  &96.55\\
 MS1M by CAST-4& 91K& 4.9M& 97.49 & 98.57 &96.52\\
\hline
 WebFace by CAST-1&2.4M &46M & 97.42 & 98.64  &97.28\\
 WebFace by CAST-2&2.1M &43M & 97.53 &  98.98 &97.51\\
 WebFace by CAST-3&2M   & 42M   & 97.65 & 99.02  &97.70\\
 WebFace by CAST-4&2M& 42M  & 97.69 & 99.08  &97.66\\
\hline
\end{tabular}}}
\end{center}

\label{tab:castcleaningresult}
\end{table}

\noindent{\bf Iterations of CAST.} Table~\ref{tab:castcleaningresult} also shows the increasing data purity after more iterations in MS1M and WebFace260M. The accuracy gradually increases from 1st to 3rd iteration, while the 4th iteration shows saturated performance. Therefore, we set the iteration number as 3 for CAST.

\noindent{\bf Performance Improvement Reporting of CAST.}
From Table~\ref{tab:castcleaningresult}, one can find the recognition performance improvement as the cleaning iteration continues. Specifically, after the first-round iteration, WebFace data results in 97.28\% TAR@FAR=1e-4 on IJB-C, which is already much higher than MS1MV2 ones (96.03\%). There are still some mislabeled identities which cause performance degradation. As the cleaning iteration continues, the number of identities decreases,
while resulting WebFace datasets further obtain 97.51\% and 97.70\% on TAR@FAR=1e-4 of IJB-C. Results on Pairs and MegaFace also lead to similar conclusions. This performance improvement reporting clearly shows the effectiveness of the proposed cleaning method.

\noindent{\bf Different Settings of Teacher/Student Model.} In Table \ref{tab:rep_1}, we further perform CAST using different settings, including ResNet-200 and MobileFaceNet backbones for Teacher/Student model, and Glint360K \cite{an2020partial} dataset for training initial Teacher model. The resulting cleaned WebFace data is then used to train face recognition models, which adopt ResNet-100 and ArcFace for fair comparisons. The experimental results are summarized as follows: (1) Replacing ResNet-100 with ResNet-200 for Teacher/Student model results in WebFace data with similar performance. We argue that ResNet-100 is competent to perform the cleaning process, and larger backbones offer marginal improvements. It is worth noting that training ResNet-200 costs several times more GPU hours than ResNet-100 ones.
(2) Replacing MS1MV2 with Glint360K for initial Teacher model also results in WebFace data with similar performance. Although Glint360K is more powerful, it is only utilized in 1st iteration of CAST, while later Teacher/Student models are all trained on WebFace data. The face recognition model trained with MS1MV2 provides a good initial embedding feature to perform 1st round cleaning, which is effective for CAST process.
(3) Replacing ResNet-100 with MobileFaceNet for Teacher/Student model shows dramatically degraded cleaning results. Due to the limited capacity, such inferior face recognition models like MobileFaceNet can not handle the ultra-large-scale WebFace training. Figure~\ref{fig:allfmr100} also illustrates the MobileFaceNet performance trained with WebFace42M, which shows very limited recognition accuracy. Considering the trade-off between cleaning cost and quality, we choose ResNet-100 for Teacher/Student model and MS1MV2 for initial Teacher model in CAST.

Furthermore, we investigate the influence of identities duplicates between MS1M and WebFace datasets. As described in Section \ref{Data_Collection}, our WebFace celebrity name list consists of two parts: the first one is borrowed from MS1M and the second one is collected from the IMDB database. Since the MS1M identities are most a subset of the WebFace ones, there are few remaining identities in MS1M if removing the duplicates.
Alternatively, we remove the MS1M identities from WebFace, which is denoted as WebFace-no-MS1M. Then, WebFace and WebFace-no-MS1M are cleaned following the same settings (the Teacher model initially uses the recognition model trained by MS1M).
We find that these 2 datasets obtain similar performance. Specifically, for IJB-C TAR@FAR=1e-4 metric, training sets cleaned from WebFace and WebFace-no-MS1M result in 97.70\% and 97.67\%, respectively. It is worth noting that MS1M name list has only 100K identities, while WebFace name list has 4M identities.
This comparison shows that duplicated identities between MS1M and WebFace do not influence the effectiveness of CAST.

    \begin{table}[!t]
\caption{Performance (\%) of different Teacher/Student models and various data for initial Teacher model . The resulting cleaned WebFace data is then used to train face recognition models, which follows the same ResNet-100 and ArcFace setting for fair comparisons.}
\begin{center}{\scalebox{0.88}{
\begin{tabular}{c|c|c|c|c}
\hline
Teacher/Student Model & \makecell[c]{Data for Initial\\Teacher Model} & Pairs & MegaFace & IJB-C \\ \hline
\hline
ResNet-100 & MS1MV2 & 97.65 & 99.02 & 97.70 \\ \hline
ResNet-200 & MS1MV2 & 97.63 & 99.07 & 97.71\\ 
ResNet-100 & Glint360K & 97.66 & 99.06 & 97.68\\  
MobileFaceNet & MS1MV2 & 97.02 & 98.24 & 95.82 \\  \hline
\end{tabular}}}
\end{center}

\label{tab:rep_1}
\end{table}

\noindent{\bf Intra-class Cleaning.} In this ablation study, we compare different intra-class cleaning modules under the framework of CAST. Both unsupervised (such as K-means \cite{kmeans} and DBSCAN \cite{ester1996density}) and supervised styles (such as GCN-D \cite{GCND} and GCN-V \cite{GCNV}) are explored to find the dominant subject in each noisy folder. As shown in Table~\ref{table:intra_cleaning}, DBSCAN achieves 96.55\% TAR@FAR=1e-4 on IJB-C, significantly outperforming K-Means (96.03\%) and slightly surpassing the supervised GCN-based ones (96.48\%  for GCN-D and 96.42\% for GCN-V). As the GCN-based strategies may be sub-optimal for the extremely noisy folders, we finally select DBSCAN \cite{ester1996density} as our intra-class cleaning module.

\begin{table}[t]
\vspace{-2mm}
\caption{Comparisons of different intra-class cleaning strategies for MS1M. ResNet-100 backbone with ArcFace loss is adopted here.}
\vspace{-2mm}
\centering
\begin{center}{\scalebox{1.0}{
\begin{tabular}{c|c|c|c|c|c}
\hline
Data & \# Id & \# Face & Pairs & MegaFace & IJB-C \\\hline\hline


 K-Means &93K &5.2M & 95.17   &97.31 & 96.03\\
 DBSCAN& 91K &4.9M & {\bf 97.42}   & {\bf 98.61}&{\bf 96.55}\\
 \hline
 GCN-D& 86K &4.4M & 96.56   & 98.55& 96.48\\
 GCN-V& 82K &4.5M & 96.93   & 98.29&96.42 \\
\hline
\end{tabular}}}
\end{center}

\label{table:intra_cleaning}
\end{table}

\begin{table}
\caption{Configuration and inference time of SFR baselines. Loose cropped test images are resized to $224 \times 224$ for joint detection and alignment. M-0.25 and R-50 refer to RetinaFace using MobileNet-0.25 (23ms) and ResNet-50 (272ms) as the backbones. FLOPs and Params mean computational complexity and parameter number of recognition module, respectively. Time refers to the duration of the whole system. For evaluation with flip setting, the batch size is set as 2.}
\centering
\begin{center}{\scalebox{0.83}{
\smallskip
\begin{tabular}{c|c|c|c|c|c}
\hline

Protocol & Det\&Align & Embedding   & FLOPs & Params & Time \\

\hline\hline
\multirow{4}{*}[-1.2ex]{\tabincell{c}{FRUITS\\-100}}
& M-0.25    & ResNet-14 & 2.1G & 19.2M& 97ms \\
& M-0.25    & MobileFaceNet (Flip) &230.3M &1.2M & 65ms \\
& M-0.25    & EfficientNet-B0 &394.2M &11.6M& 94ms\\
& M-0.25    & RegNet-800MF &831.0M &23.4M& 89ms \\ \hline

\multirow{4}{*}[-1.2ex]{\tabincell{c}{FRUITS\\-500}}
& R-50    & ResNet-100  &12.1G &65.2M & 481ms \\
& R-50    & ResNet-50 (Flip)  &6.3G & 43.6M& 492ms  \\
& R-50    & SENet-50  &6.3G & 43.8M&  374ms \\
& R-50    & ResNeXt-100  &8.2G &56.2M & 411ms  \\
& R-50    & RegNet-8GF  & 8.0G&82.7M& 429ms \\ \hline

\multirow{5}{*}[-1.2ex]{\tabincell{c}{FRUITS\\-1000}}
& R-50    & ResNet-100 (Flip)  & 12.1G&65.2M & 826ms \\
& R-50    & ResNet-200  &23.9G &109.3M & 892ms \\
& R-50    & SENet-152  & 18.1G& 101.0M& 792ms \\
& R-50    & AttentionNet-152  & 14.8G&61.3M & 785ms \\
& R-50    & RegNet-16GF &16.0G &103.7M&  772ms \\ \hline

\hline
\end{tabular}}}
\end{center}

\label{table:time}
\end{table}

\subsection{Baselines under FRUITS Protocol}
\label{sec:FRUITS-100}

\subsubsection{Standard Face Recognition}

In this section, we set up a series of SFR baselines under the proposed FRUITS protocol. Table~\ref{table:time} illustrates various face recognition systems (including different settings of detection, alignment, feature embedding) and their inference time. In our SFR baselines, representative network architectures are explored, covering MobileNet \cite{MobileNet,chen2018mobilefacenets}, EfficientNet \cite{tan2019efficientnet}, AttentionNet \cite{AttentionNet}, ResNet \cite{ResNet}, SENet \cite{SENet}, ResNeXt \cite{ResNeXt} and RegNet \cite{RegNet} families. All the models are trained on WebFace42M with ArcFace.

\begin{figure*}
\small
\centering

\subfigure[{\scriptsize FMR-FNMR for FRUITS-100}]{
\label{fig:allfmr100}
\includegraphics[width=0.3\linewidth]{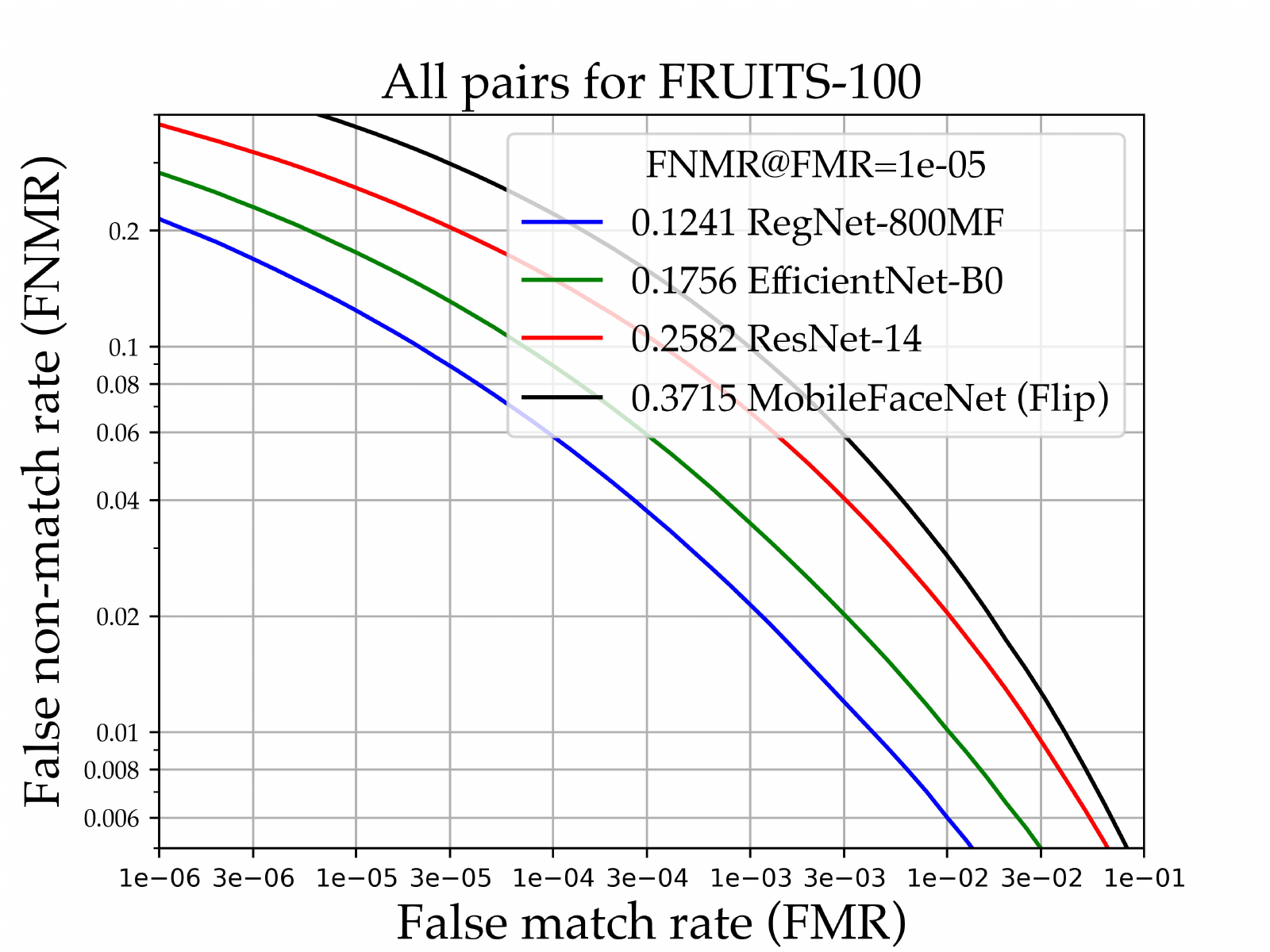}}
\subfigure[{\scriptsize FMR-FNMR for FRUITS-500}]{
\label{fig:allfmr500}
\includegraphics[width=0.3\linewidth]{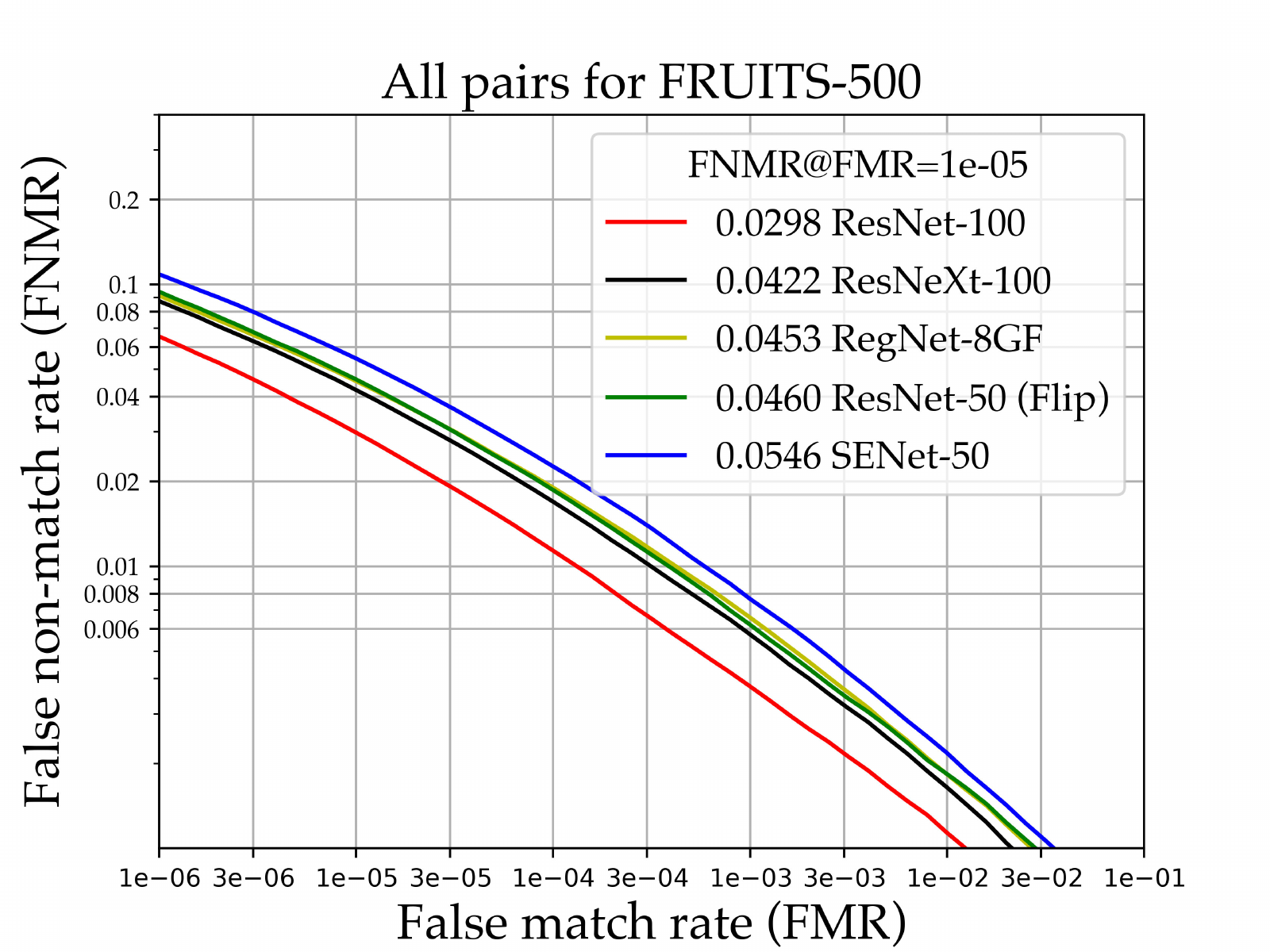}}
\subfigure[{\scriptsize FMR-FNMR for FRUITS-1000}]{
\label{fig:allfmr1000}
\includegraphics[width=0.3\linewidth]{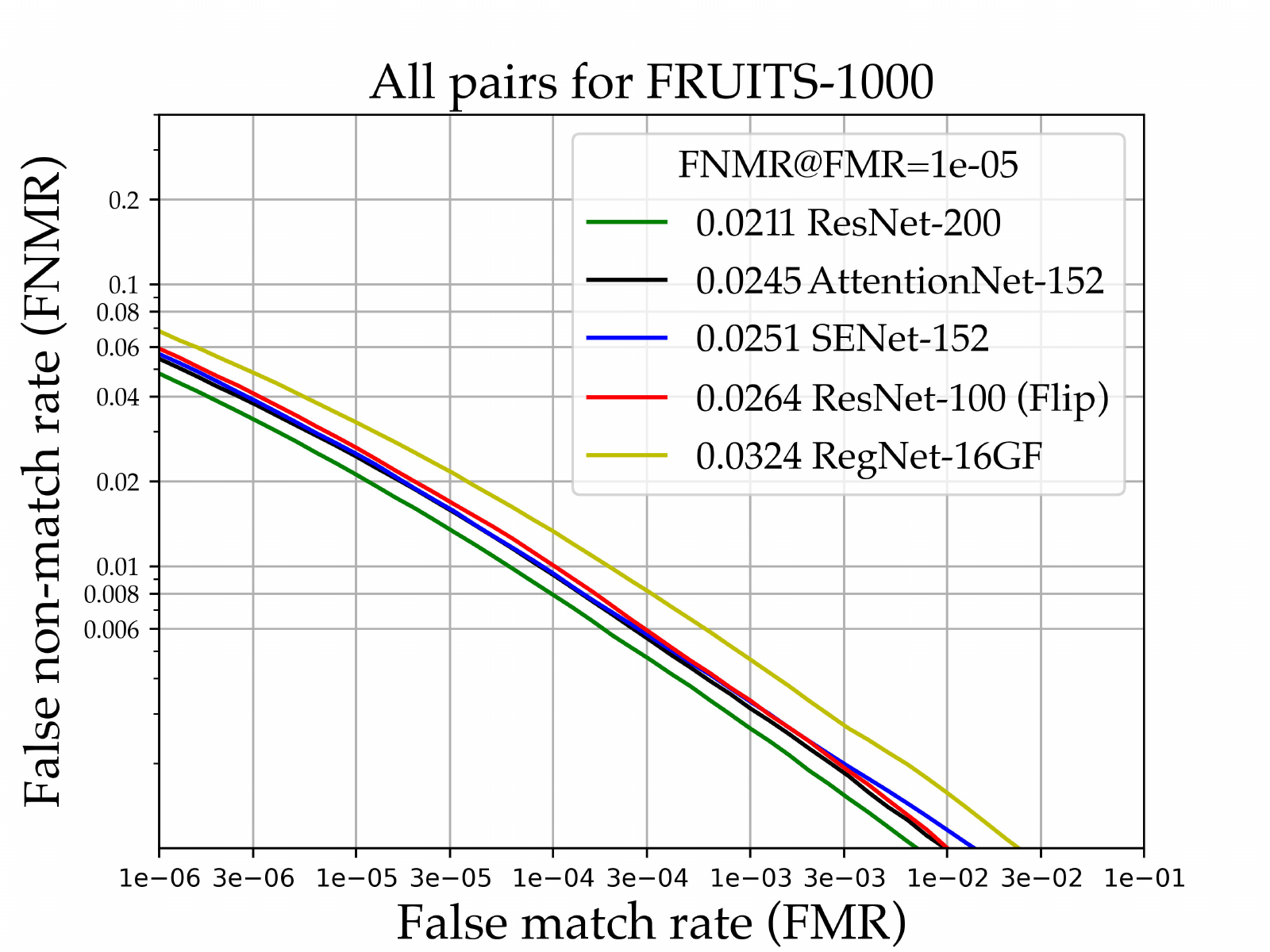}}

\caption{Performance comparisons of SFR under the FRUITS. The FMR-FNMR plots for \emph{All} pairs verification are drawn, and models are ranked in legend according to FNMR@FMR=1e-5 (\textbf{lower FNMR is better}).}
\label{fig:fruits}
\end{figure*}

\begin{figure*}
\small
\centering

\subfigure[{\scriptsize Attributes for FRUITS-100}]{
\label{fig:allattribute100}
\includegraphics[width=0.3\linewidth]{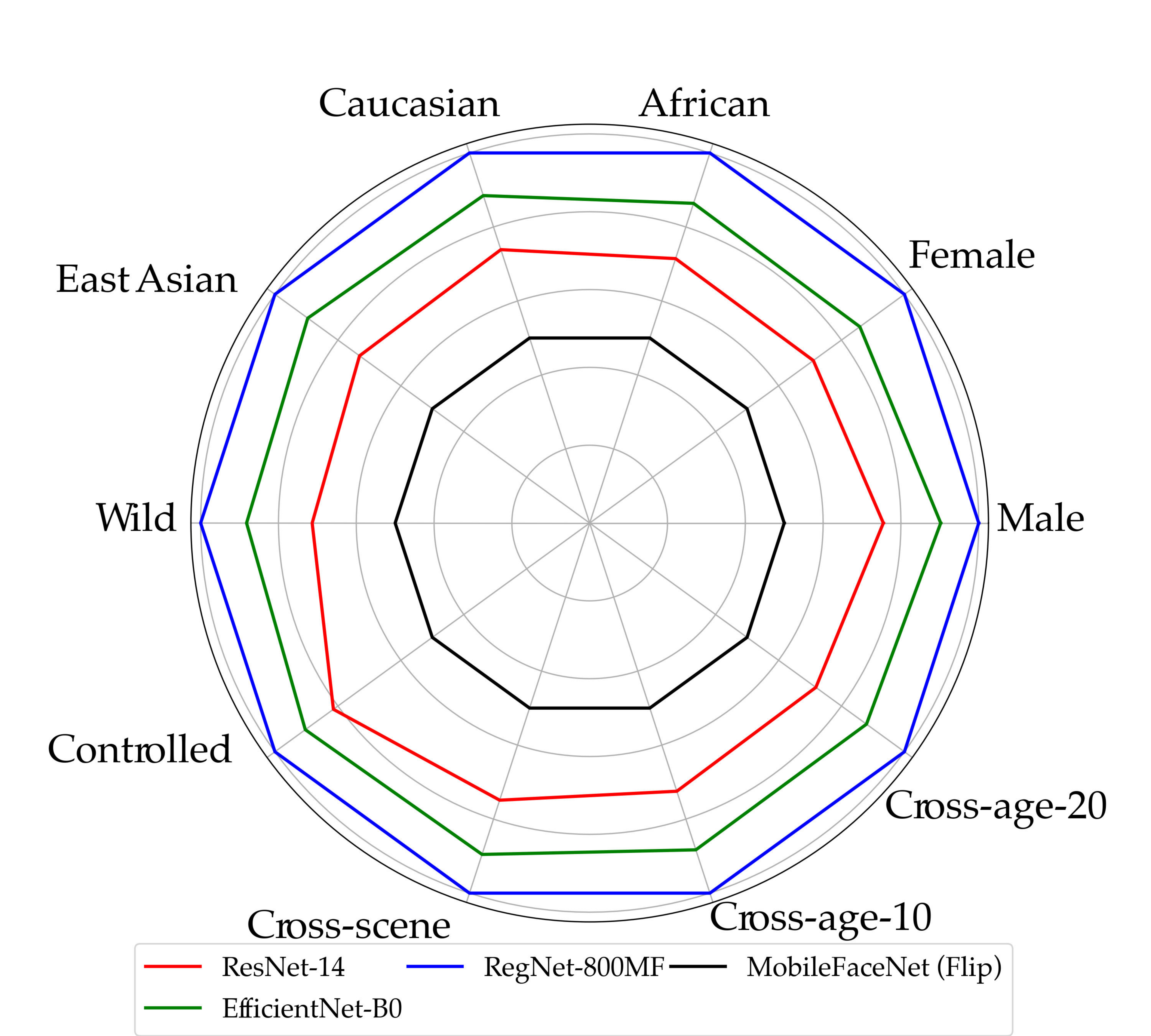}}
\subfigure[{\scriptsize Attributes for FRUITS-500}]{
\label{fig:allattribute500}
\includegraphics[width=0.3\linewidth]{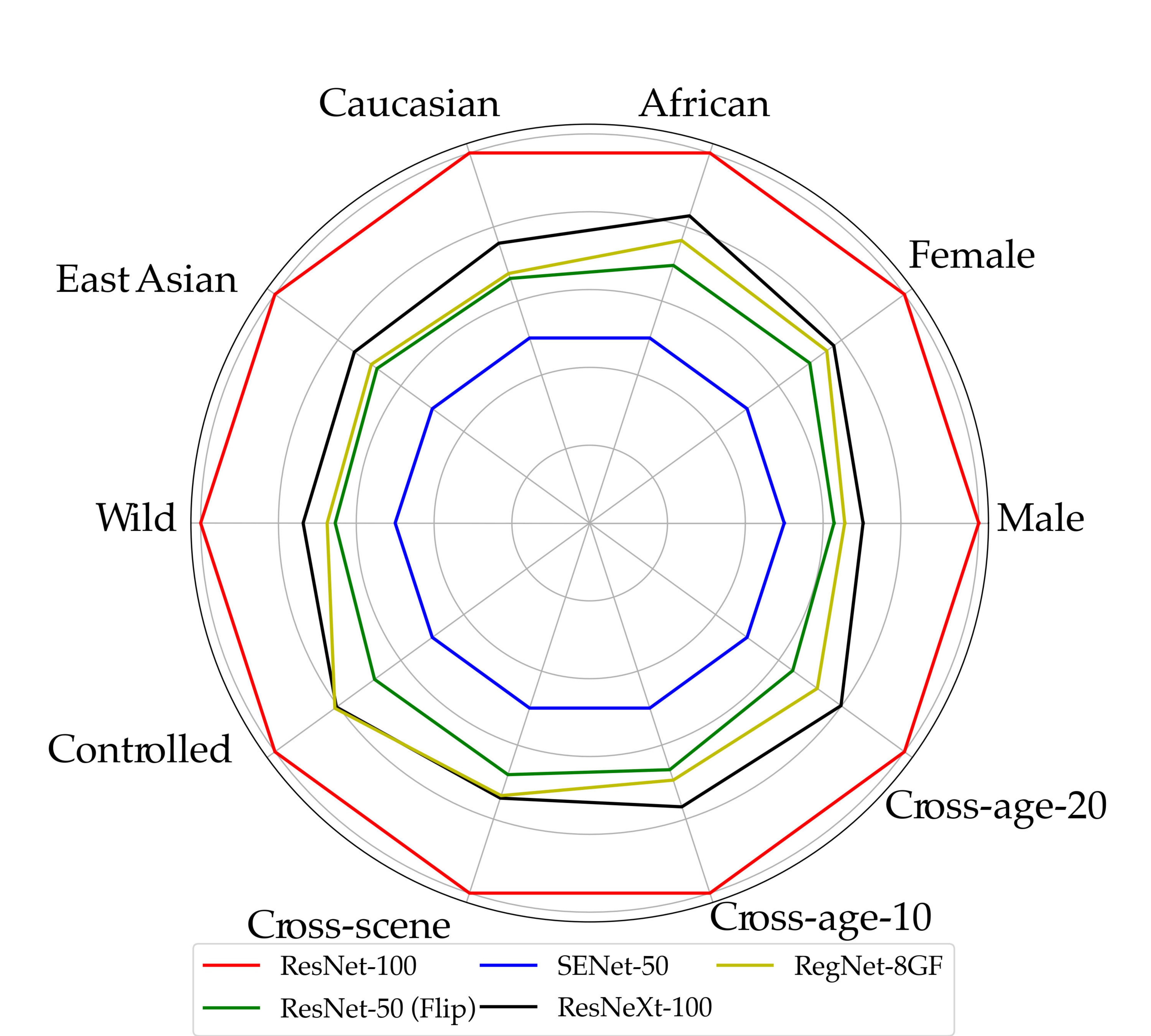}}
\subfigure[{\scriptsize Attributes for FRUITS-1000}]{
\label{fig:allattribute1000}
\includegraphics[width=0.3\linewidth]{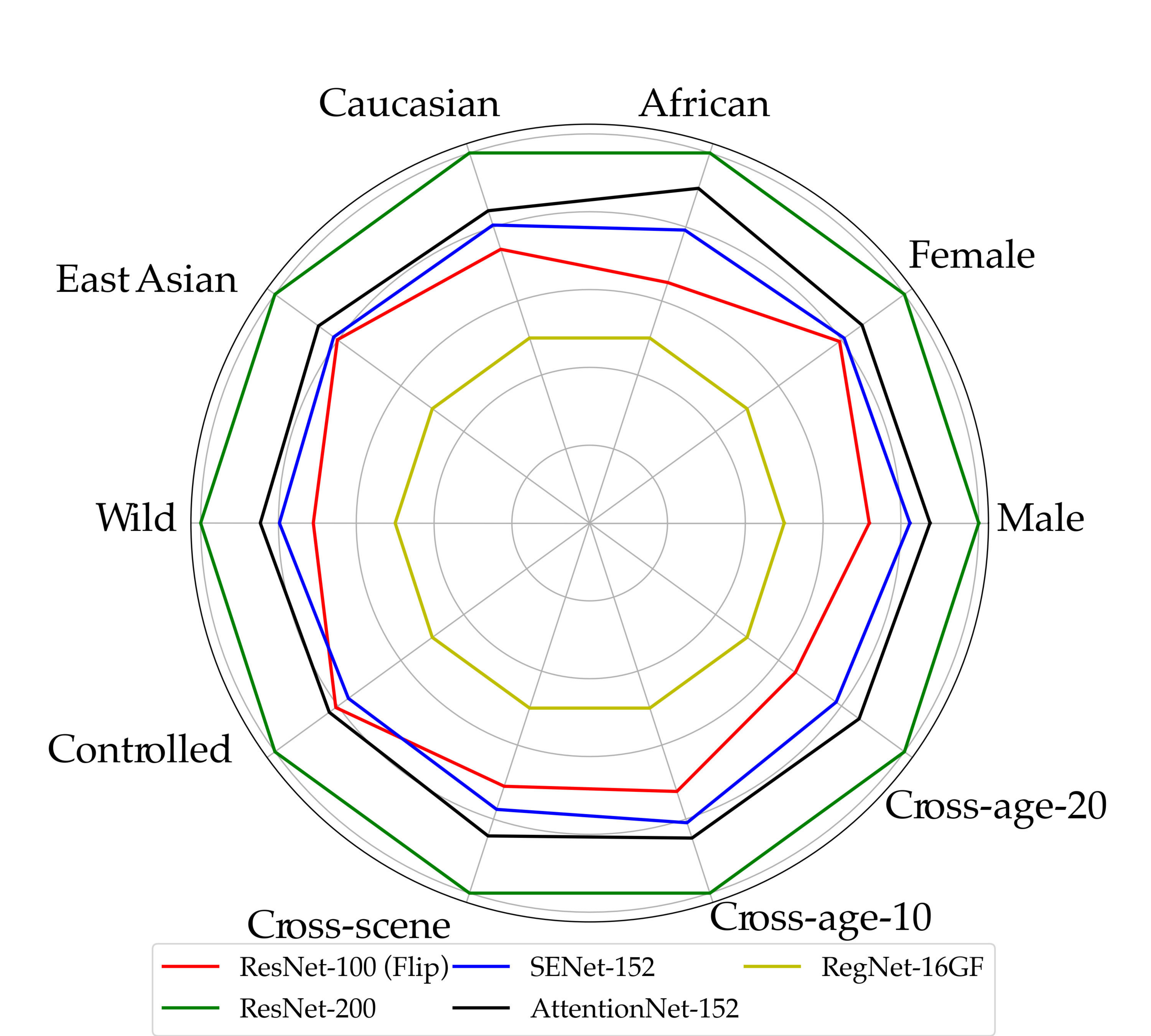}}
\caption{Attributes ranking of different models under the FRUITS. We show the attribute plots under FNMR@FMR=1e-5, which is normalized to 0.5-1.0 for better visualization (\textbf{outer is better}).}
\label{fig:fruits_attributes}
\end{figure*}

Due to the strict time limitation, FRUITS-100 track can only adopt lightweight architectures, including RetinaFace-MobileNet-0.25 \cite{RetinaFace} for face detection and alignment, ResNet-14, MobileFaceNet (Flip), EfficientNet-B0 and RegNet-800MF for face feature extraction. FMR-FNMR plots on \emph{All} pairs and analyses of attributes are shown in Figure~\ref{fig:allfmr100} and Figure~\ref{fig:allattribute100}. Because of the weak detection and recognition modules, the best baseline (RegNet-800MF) only obtains 12.41\% FNMR@FMR=1e-5 (lower is better). Therefore, there leaves substantial room for future improvement under the FRUITS-100 protocol.

For the FRUITS-500 protocol, we can employ more capable modern networks, such as RetinaFace-ResNet-50 \cite{RetinaFace} for pre-processing, and ResNet-100, ResNet-50 (Flip), SENet-50, ResNeXt-100, RegNet-8GF for feature embedding extraction. As shown in Figure~\ref{fig:allfmr500} and Figure~\ref{fig:allattribute500}, ResNet-100 exhibits the best overall performance, scoring 2.98\% FNMR@FMR=1e-5.
For attributes evaluation, ResNet-100 also achieves the lowest FNMR according to the indicators of age, scenarios, race and gender.

Recognition models under the FRUITS-1000 protocol can be more complicated and powerful, therefore we explore ResNet-100 (Flip), ResNet-200, SENet-152, AttentionNet-152 and RegNet-16GF for face feature representations. As shown in Figure~\ref{fig:allfmr1000} and Figure~\ref{fig:allattribute1000}, ResNet-200 performs best in face verification and wins all attribute comparisons. Compared with lightweight FRUITS-100 track, performances of different large models are much closer. This result implies that new designs need to be explored for the heavyweight FRUITS track.

\subsubsection{Masked Face Recognition}

Recognizing identities with masks may be the most challenging face recognition problem, which is
essential for biometric authentication during COVID-19. Based on the proposed WebFace benchmark, we perform MFR in this part, and establish a series of baselines for different training settings. ResNet-100 in FRUITS-500 protocol and ArcFace are adopted, while similar conclusions can be drawn for other backbones or losses.

As shown in Table \ref{table:masked_face_recognition}, the best-performed public training set, MS1MV2, only scores 71.81\% FNMR@FMR=10-5 for \emph{All-Masked} comparisons on our test set (with ArcFace loss). Different proportions of the WebFace data reduce FNMR to 70.97\%, 56.47\% and 47.25\% respectively, which shows the superiority of the proposed dataset again.
Moreover, we augment the training data with simulated masks to further investigate this difficult recognition scenario. Specifically, mask renderer in \cite{insightface_mask_renderer} is applied on each face, where wearing height and mask types are randomly chosen. Both simulated and original faces are trained together with 20 epochs. In Table \ref{table:masked_face_recognition}, one can find that this simple augmentation strategy effectively boosts the MFR accuracy. For all data proportions, the FNMR improvements are near 10\%, which build strong baselines for future MFR researches.

\begin{table}[!t]
\caption{Performance (\%) of different training data for MFR, including \emph{Controlled-Masked}, \emph{Wild-Masked}, and \emph{All-Masked} comparisons. \emph{Mask} suffix of training data refers to augmentation with simulated masks. FNMR@FMR=10-5 of ResNet-100 and ArcFace is reported.}
\begin{center}{\scalebox{0.88}{
\begin{tabular}{c|c|c|c}
\hline
Data & Controlled-Masked & Wild-Masked & All-Masked \\ \hline
\hline
MS1MV2 & 64.40 & 76.07 & 71.81  \\ \hline
WebFace4M & 61.48 & 75.43 &  70.97 \\ 
WebFace4M-Mask & 49.58 & 65.53 & 59.89  \\  \hline
WebFace12M & 45.46 & 61.57 & 56.47  \\  
WebFace12M-Mask & 31.61 & 46.98 & 41.45  \\  \hline
WebFace42M & 36.81 & 52.83 & 47.25  \\  
WebFace42M-Mask & \textbf{24.60} & \textbf{39.42} & \textbf{33.87}  \\  \hline

\end{tabular}}}
\end{center}

\label{table:masked_face_recognition}
\end{table}

\begin{table*}[t]
\caption{Performance of different training data for UFR. For race and gender groups, FNMR@FMR=10-5 of ResNet-100 and ArcFace is reported.
STD and SER are main fairness evaluation metrics.}
\centering
\begin{center}{\scalebox{1.0}{
\begin{tabular}{c|c|c|c|c|c|c|c|c|c|c|c}
\hline
\multirow{2}{*}{Data} &\multicolumn{6}{c|}{Race} & \multicolumn{5}{c}{Gender} \\

\cline{2-12}

& Caucasian  & East Asian & African  & Avg & \textbf{STD} & \textbf{SER} & Male & Famale & Avg & \textbf{STD} & \textbf{SER} \\



\hline \hline

MS1MV2

& 0.1050 & 0.1474 & 0.1053 & 0.1192 & 0.0199 & 1.40&  0.0850 & 0.1597 & 0.1224 & 0.0374  & 1.88 \\

 \hline

WebFace4M

& 0.0942 & 0.1368 & 0.1071 & 0.1127 & 0.0178 & 1.45 & 0.0890 & 0.1713 & 0.1302  & 0.0412  & 1.92 \\








WebFace4M-Balanced

& 0.0976 & 0.1253 & 0.1025 & 0.1085 &  \textbf{0.0121} & \textbf{1.28}
& 0.0825 & 0.1332 & 0.1079 & \textbf{0.0254} & \textbf{1.61}\\

\hline
\end{tabular}}}
\end{center}

\label{tab:performance_UFR}
\end{table*}

\begin{table*}[!t]
\caption{Results on NIST-FRVT.  Numbers in bracket are rankings of different tracks.
Our ResNet-200 model using Arcface is trained on WebFace42M. FNMR at corresponding FMR is reported.}
\begin{center}{\scalebox{0.92}{
\begin{tabular}{c|c|c|c|c|c|c|c}
\hline
Rank & entries & \makecell[c]{\textbf{Visa}\\\scriptsize{FNMR@FMR=10-6}} & \makecell[c]{\textbf{Mugshot}\\\scriptsize{FNMR@FMR=10-5}} & \makecell[c]{\textbf{Mugshot DT$\geq$12Years} \\ \scriptsize{FNMR@FMR=10-5}} & \makecell[c]{\textbf{VisaBorder}\\\scriptsize{FNMR@FMR=10-6}}  & \makecell[c]{\textbf{Border}\\\scriptsize{FNMR@FMR=10-6}}  &\makecell[c]{\textbf{Wild}\\\scriptsize{FNMR@FMR=10-5}} \\ \hline
\hline
1 & deepglint & 0.0027 (3) & 0.0032 (8) &0.0033 (4)& 0.0043 (3) & 0.0084 (4) &  0.0301 (4) \\
2 & visionlabs &0.0025 (1)  & 0.0026 (1) &0.0029 (3)& 0.0035 (1) &  0.0064 (1)&  0.0306 (13) \\ \hline
3 & \textbf{ours} & \textbf{0.0034 (5)} & \textbf{0.0027 (3)} & \textbf{0.0028 (2)} &\textbf{0.0046 (6)}  & \textbf{0.0088 (6)} &  \textbf{0.0303 (10)} \\ \hline
4 & dahua & 0.0046 (13) & 0.0035 (12) & 0.0049 (13)&0.0046 (5) & 0.0076 (2) &  0.0300 (3) \\
 5 & cib & 0.0061 (20) & 0.0030 (6) & 0.0041 (7) &0.0048 (8) & 0.0578 (76) &  0.0302 (25) \\
 6 & nazhi & 0.0059 (16) & 0.0036  (13)& 0.0048 (12)&0.0057 (12) & 0.0125 (19) &  0.0300 (2) \\
 7 & vocord & 0.0038 (6) & 0.0042 (18) & 0.0055 (15)&0.0045 (4) & 0.0086 (5) &  0.0310 (24) \\
  8 & ercacat & 0.0044 (10) & 0.0033 (9) & 0.0047 (10)&0.0106 (43) & 0.0202 (36) &  0.0293 (1) \\
   9 & everai-paravision & 0.0050 (14) & 0.0036 (14) & 0.0052 (14)&0.0092 (35) & 0.0193 (35) &  0.0302 (7) \\
    10 & aimall & 0.0041 (8) & 0.0033 (10) & 0.0035 (5)&0.0056 (11)& 0.0109 (15) &  0.0335 (49) \\

\hline
\end{tabular}}}
\end{center}

\label{table:nist}
\vspace{-0.4cm}
\end{table*}

\subsubsection{Unbiased Face Recognition}
Generalized deployments call for robust and fair face recognition systems. In this part, we perform data sampling on WebFace data and investigate its fairness influence on our test set. Table \ref{tab:performance_UFR} indicates that
MS1MV2 and WebFace4M show considerable bias among race and gender. The scores of SER metric are 1.40, 1.45, 1.88, and 1.92 respectively.
The relative rank of different models (trained with WebFace42M) on race and gender attributes is also illustrated in Figure~\ref{fig:fruits_attributes}.
Thanks to the ultra-large-scale of the proposed benchmark, we can sample a balanced race/gender sub-set denoted as WebFace4M-Balanced.
According to STD and SER scores, this sampled training data reduces the recognition bias to some extent, surpassing the MS1MV2 and  WebFace4M.
It is worth noting that there is still demographic bias on the test set even with WebFace4M-Balanced training, which is consistent with the observation in previous studies \cite{RFW, wang2019mitigate}. More bias-mitigating solutions need to be developed such as loss design, augmentation, and adversarial learning.


In summary, the results show the great potential of WebFace data (including training and test set) for more fair and robust face recognition systems. Based on the proposed WebFace benchmark, we hope to spark UFR researches in the future.

\subsection{Results on NIST-FRVT}
Finally, we report the submission to NIST-FRVT. Following the settings of FRUITS-1000, our system is built based on RetinaFace-ResNet-50 for detection and alignment, and ArcFace-ResNet-200 trained on WebFace42M for feature embedding extraction. The network is accelerated by OpenVINO~\cite{openvino} and the flip test is adopted. The final inference time is near 1300 milliseconds according to the NIST-FRVT report, meeting the latest 1500 milliseconds limitation. Table~\ref{table:nist} illustrates top-ranking entries measured by FNMR across six tracks \footnote{According to report of October 9, 2020}. \textbf{Our model trained on the WebFace42M obtains overall 3rd among 430 submissions}, showing impressive performance across different tracks.
Specifically, the proposed solution based on WebFace ranks 5th and 3rd on controlled \emph{Visa} and \emph{Mugshot} scenarios, respectively. On more challenging cross-age comparisons (\emph{Mugshot DT$\geq$12Years}), we get 2nd place. For less-controlled \emph{VisaBorder}, \emph{Border} and \emph{Wild} tracks, state-of-the-art performances are also achieved.
 Considering hundreds of company entries to NIST-FRVT, the WebFace42M takes a significant step towards closing the data gap between academia and industry.

\section{Discussion and Conclusion}
\label{sec:discussion}

\noindent{\bf{Discussion}} During WebFace260M dataset construction, privacy and bias issues are our primary concerns. For privacy protection, all face images including training and test data are collected from public Internet resources. For data download, we provide strict access for qualified research groups that sign the license, and try our best to guarantee WebFace260M for research purposes only. For bias problem, our dataset has diverse birth dates, poses and ages, while gender and race are inevitably biased due to complex nationality and profession distributions. In evaluations of this work, we
especially design the Unbiased Face Recognition (UFR), studying the influence of balanced training data. We argue that the community could develop more bias-mitigating solutions based on ultra-large-scale WebFace260M benchmark.

\vspace{0.1cm}
\noindent{\bf{Conclusion}} In this paper, we have dived into the million-scale face recognition problem, contributing a tremendous noisy dataset with 260M faces,
 a high-quality training dataset with 42M images of 2M identities by using automatic cleaning, a test set containing rich attributes and large-scale masked face sub-set, a time-constrained evaluation protocol, a distributed framework at linear acceleration, a succession of baselines on various scenarios, as well as a final state-of-the-art model. Equipped with this publicly available face dataset, our model significantly reduces 40\% failure rate on IJB-C and ranks 3rd among 430 entries on NIST-FRVT. Besides, baselines built on the proposed WebFace show great potential for masked and unbiased recognition tasks. We hope this benchmark could close the data gap behind the industry, and facilitate future researches of ultra-large-scale face recognition.


\small
\bibliographystyle{ieee}

\bibliography{egbib}

\begin{thebibliography}{100}\itemsep=-1pt

\bibitem{glintweb}
\url{http://trillionpairs.deepglint.com/overview}.

\bibitem{FRVT}
\url{https://www.nist.gov/programs-projects/face-recognition-vendor-test-frvt-ongoing}.

\bibitem{bing_image}
\url{https://www.bing.com/image}.

\bibitem{freebase}
\url{https://developers.google.com/freebase/}.

\bibitem{imdb_website}
\url{https://www.imdb.com/}.

\bibitem{google_image}
\url{https://images.google.com}.

\bibitem{insightface_mask_renderer}
\url{https://github.com/deepinsight/insightface/tree/master/recognition/_tools_/}.

\bibitem{openvino}
\url{https://github.com/openvinotoolkit/openvino}.

\bibitem{an2020partial}
X.~An, X.~Zhu, Y.~Xiao, L.~Wu, M.~Zhang, Y.~Gao, B.~Qin, D.~Zhang, and Y.~Fu.
\newblock Partial {FC}: Training 10 million identities on a single machine.
\newblock {\em arXiv:2010.05222}, 2020.

\bibitem{anwar2020masked}
A.~Anwar and A.~Raychowdhury.
\newblock Masked face recognition for secure authentication.
\newblock {\em arXiv:2008.11104}, 2020.

\bibitem{UMDFaces}
A.~Bansal, A.~Nanduri, C.~D. Castillo, R.~Ranjan, and R.~Chellappa.
\newblock {UMDF}aces: An annotated face dataset for training deep networks.
\newblock {\em arXiv:1611.01484v2}, 2016.

\bibitem{IJCB-mask}
F.~Boutros, N.~Damer, J.~N. Kolf, K.~Raja, F.~Kirchbuchner, R.~Ramachandra,
  A.~Kuijper, P.~Fang, C.~Zhang, F.~Wang, et~al.
\newblock {MFR} 2021: Masked face recognition competition.
\newblock {\em IJCB Workshop}, 2021.

\bibitem{cao2020domain}
D.~Cao, X.~Zhu, X.~Huang, J.~Guo, and Z.~Lei.
\newblock Domain balancing: Face recognition on long-tailed domains.
\newblock In {\em CVPR}, 2020.

\bibitem{VGGFace2}
Q.~Cao, L.~Shen, W.~Xie, O.~M. Parkhi, and A.~Zisserman.
\newblock {VGGF}ace2: A dataset for recognising faces across pose and age.
\newblock In {\em FG}, 2018.

\bibitem{chen2018mobilefacenets}
S.~Chen, Y.~Liu, X.~Gao, and Z.~Han.
\newblock Mobile{F}acenets: Efficient {CNNs} for accurate real-time face
  verification on mobile devices.
\newblock In {\em CCBR}, 2018.

\bibitem{chopra2005learning}
S.~Chopra, R.~Hadsell, and Y.~LeCun.
\newblock Learning a similarity metric discriminatively, with application to
  face verification.
\newblock In {\em CVPR}, 2005.

\bibitem{damer2021extended}
N.~Damer, F.~Boutros, M.~S{\"u}{\ss}milch, F.~Kirchbuchner, and A.~Kuijper.
\newblock Extended evaluation of the effect of real and simulated masks on face
  recognition performance.
\newblock {\em IET Biometrics}, 2021.

\bibitem{deng2021masked}
J.~Deng, J.~Guo, X.~An, Z.~Zhu, and S.~Zafeiriou.
\newblock Masked face recognition challenge: The {InsightFace} track report.
\newblock {\em arXiv preprint arXiv:2108.08191}, 2021.

\bibitem{subcenter}
J.~Deng, J.~Guo, T.~Liu, M.~Gong, and S.~Zafeiriou.
\newblock Sub-center {ArcFace}: Boosting face recognition by large-scale noisy
  web faces.
\newblock In {\em ECCV}, 2020.

\bibitem{RetinaFace}
J.~Deng, J.~Guo, E.~Ververas, I.~Kotsia, and S.~Zafeiriou.
\newblock {RetinaFace}: Single-shot multi-level face localisation in the wild.
\newblock In {\em CVPR}, 2020.

\bibitem{ArcFace}
J.~Deng, J.~Guo, and S.~Zafeiriou.
\newblock Arc{F}ace: Additive angular margin loss for deep face recognition.
\newblock In {\em CVPR}, 2019.

\bibitem{LFR}
J.~Deng, J.~Guo, D.~Zhang, Y.~Deng, X.~Lu, and S.~Shi.
\newblock Lightweight face recognition challenge.
\newblock In {\em ICCV Workshop}, 2019.

\bibitem{deng2017marginal}
J.~Deng, Y.~Zhou, and S.~Zafeiriou.
\newblock Marginal loss for deep face recognition.
\newblock In {\em CVPR Workshop}, 2017.

\bibitem{ding2020masked}
F.~Ding, P.~Peng, Y.~Huang, M.~Geng, and Y.~Tian.
\newblock Masked face recognition with latent part detection.
\newblock In {\em ACM MM}, 2020.

\bibitem{du2020semi}
H.~Du, H.~Shi, Y.~Liu, J.~Wang, Z.~Lei, D.~Zeng, and T.~Mei.
\newblock Semi-siamese training for shallow face learning.
\newblock In {\em ECCV}, 2020.

\bibitem{du2021towards}
H.~Du, H.~Shi, Y.~Liu, D.~Zeng, and T.~Mei.
\newblock Towards nir-vis masked face recognition.
\newblock {\em IEEE SPL}, 2021.

\bibitem{duan2019uniformface}
Y.~Duan, J.~Lu, and J.~Zhou.
\newblock Uniform{F}ace: Learning deep equidistributed representation for face
  recognition.
\newblock In {\em CVPR}, 2019.

\bibitem{eikenberry2020mask}
S.~E. Eikenberry, M.~Mancuso, E.~Iboi, T.~Phan, K.~Eikenberry, Y.~Kuang,
  E.~Kostelich, and A.~B. Gumel.
\newblock To mask or not to mask: Modeling the potential for face mask use by
  the general public to curtail the {COVID}-19 pandemic.
\newblock {\em Infectious Disease Modelling}, 2020.

\bibitem{ester1996density}
M.~Ester, H.-P. Kriegel, J.~Sander, X.~Xu, et~al.
\newblock A density-based algorithm for discovering clusters in large spatial
  databases with noise.
\newblock In {\em ACM KDD}, 1996.

\bibitem{geng2020masked}
M.~Geng, P.~Peng, Y.~Huang, and Y.~Tian.
\newblock Masked face recognition with generative data augmentation and domain
  constrained ranking.
\newblock In {\em ACM MM}, 2020.

\bibitem{gong2020jointly}
S.~Gong, X.~Liu, and A.~K. Jain.
\newblock Jointly de-biasing face recognition and demographic attribute
  estimation.
\newblock In {\em ECCV}, 2020.

\bibitem{gong2021mitigating}
S.~Gong, X.~Liu, and A.~K. Jain.
\newblock Mitigating face recognition bias via group adaptive classifier.
\newblock In {\em CVPR}, 2021.

\bibitem{goyal2017accurate}
P.~Goyal, P.~Doll{\'a}r, R.~Girshick, P.~Noordhuis, L.~Wesolowski, A.~Kyrola,
  A.~Tulloch, Y.~Jia, and K.~He.
\newblock Accurate, large minibatch {SGD}: Training imagenet in 1 hour.
\newblock {\em arXiv:1706.02677}, 2017.

\bibitem{grother2019face}
P.~Grother, M.~Ngan, and K.~Hanaoka.
\newblock {\em Face Recognition Vendor Test (FVRT): Part 3, Demographic
  Effects}.
\newblock National Institute of Standards and Technology, 2019.

\bibitem{MS1M}
Y.~Guo, L.~Zhang, Y.~Hu, X.~He, and J.~Gao.
\newblock {MS-Celeb-1M}: A dataset and benchmark for large-scale face
  recognition.
\newblock In {\em ECCV}, 2016.

\bibitem{hariri2021efficient}
W.~Hariri.
\newblock Efficient masked face recognition method during the {COVID}-19
  pandemic.
\newblock {\em arXiv:2105.03026}, 2021.

\bibitem{ResNet}
K.~He, X.~Zhang, S.~Ren, and J.~Sun.
\newblock Deep residual learning for image recognition.
\newblock In {\em CVPR}, 2016.

\bibitem{MobileNet}
A.~G. Howard, M.~Zhu, B.~Chen, D.~Kalenichenko, W.~Wang, T.~Weyand,
  M.~Andreetto, and H.~Adam.
\newblock Mobilenets: Efficient convolutional neural networks for mobile vision
  applications.
\newblock {\em arXiv:1704.04861}, 2017.

\bibitem{SENet}
J.~Hu, L.~Shen, and G.~Sun.
\newblock Squeeze-and-excitation networks.
\newblock In {\em CVPR}, 2018.

\bibitem{hu2019noise}
W.~Hu, Y.~Huang, F.~Zhang, and R.~Li.
\newblock Noise-tolerant paradigm for training face recognition cnns.
\newblock In {\em CVPR}, 2019.

\bibitem{LFW}
G.~B. Huang, M.~Ramesh, T.~Berg, and E.~Learned-Miller.
\newblock Labeled faces in the wild: A database for studying face recognition
  in unconstrained environments.
\newblock Technical report, 2007.

\bibitem{Curricularface}
Y.~Huang, Y.~Wang, Y.~Tai, X.~Liu, P.~Shen, S.~Li, J.~Li, and F.~Huang.
\newblock Curricular{F}ace: adaptive curriculum learning loss for deep face
  recognition.
\newblock In {\em CVPR}, 2020.

\bibitem{huang2021age}
Z.~Huang, J.~Zhang, and H.~Shan.
\newblock When age-invariant face recognition meets face age synthesis: A
  multi-task learning framework.
\newblock In {\em CVPR}, 2021.

\bibitem{MegaFace}
I.~Kemelmacher-Shlizerman, S.~M. Seitz, D.~Miller, and E.~Brossard.
\newblock The {MegaFace} benchmark: 1 million faces for recognition at scale.
\newblock In {\em CVPR}, 2016.

\bibitem{Groupface}
Y.~Kim, W.~Park, M.-C. Roh, and J.~Shin.
\newblock Group{F}ace: Learning latent groups and constructing group-based
  representations for face recognition.
\newblock In {\em CVPR}, 2020.

\bibitem{BroadFace}
Y.~Kim, W.~Park, and J.~Shin.
\newblock Broad{F}ace: Looking at tens of thousands of people at once for face
  recognition.
\newblock In {\em ECCV}, 2020.

\bibitem{AlexNet}
A.~Krizhevsky, I.~Sutskever, and G.~E. Hinton.
\newblock {ImageNet} classification with deep convolutional neural networks.
\newblock In {\em NeurIPS}, 2012.

\bibitem{kwon2021association}
S.~Kwon, A.~D. Joshi, C.-H. Lo, D.~A. Drew, L.~H. Nguyen, C.-G. Guo, W.~Ma,
  R.~S. Mehta, F.~M. Shebl, E.~T. Warner, et~al.
\newblock Association of social distancing and face mask use with risk of
  covid-19.
\newblock {\em Nature Communications}, 2021.

\bibitem{li2021dynamic}
B.~Li, T.~Xi, G.~Zhang, H.~Feng, J.~Han, J.~Liu, E.~Ding, and W.~Liu.
\newblock Dynamic class queue for large scale face recognition in the wild.
\newblock In {\em CVPR}, 2021.

\bibitem{li2021virtual}
P.~Li, B.~Wang, and L.~Zhang.
\newblock Virtual fully-connected layer: Training a large-scale face
  recognition dataset with limited computational resources.
\newblock In {\em CVPR}, 2021.

\bibitem{liu2019fair}
B.~Liu, W.~Deng, Y.~Zhong, M.~Wang, J.~Hu, X.~Tao, and Y.~Huang.
\newblock Fair loss: Margin-aware reinforcement learning for deep face
  recognition.
\newblock In {\em ICCV}, 2019.

\bibitem{liu2019adaptiveface}
H.~Liu, X.~Zhu, Z.~Lei, and S.~Z. Li.
\newblock Adaptiveface: Adaptive margin and sampling for face recognition.
\newblock In {\em CVPR}, 2019.

\bibitem{A-SoftMax}
W.~Liu, Y.~Wen, Z.~Yu, M.~Li, B.~Raj, and L.~Song.
\newblock Sphereface: Deep hypersphere embedding for face recognition.
\newblock In {\em CVPR}, 2017.

\bibitem{IQIYI2018}
Y.~Liu, P.~Shi, B.~Peng, H.~Yan, Y.~Zhou, B.~Han, Y.~Zheng, C.~Lin, J.~Jiang,
  and Y.~Fan.
\newblock {IQIYI-VID}: A large dataset for multi-modal person identification.
\newblock {\em arXiv:1811.07548}, 2018.

\bibitem{kmeans}
S.~Lloyd.
\newblock Least squares quantization in {PCM}.
\newblock {\em TIT}, 1982.

\bibitem{IJB-C}
B.~Maze, J.~Adams, J.~A. Duncan, N.~Kalka, T.~Miller, C.~Otto, A.~K. Jain,
  W.~T. Niggel, J.~Anderson, and J.~Cheney.
\newblock {IARPA Janus Benchmark C}: Face dataset and protocol.
\newblock In {\em ICB}, 2018.

\bibitem{AgeDB}
S.~Moschoglou, A.~Papaioannou, C.~Sagonas, J.~Deng, I.~Kotsia, and
  S.~Zafeiriou.
\newblock {AgeDB}: The first manually collected in-the-wild age database.
\newblock In {\em CVPR Workshop}, 2017.

\bibitem{MF2}
A.~Nech and I.~Kemelmacher-Shlizerman.
\newblock Level playing field for million scale face recognition.
\newblock In {\em CVPR}, 2017.

\bibitem{negi2021deep}
A.~Negi, K.~Kumar, P.~Chauhan, and R.~Rajput.
\newblock Deep neural architecture for face mask detection on simulated masked
  face dataset against {COVID}-19 pandemic.
\newblock In {\em ICCIS}, 2021.

\bibitem{FaceScrub}
H.-W. Ng and S.~Winkler.
\newblock A data-driven approach to cleaning large face datasets.
\newblock In {\em ICIP}, 2014.

\bibitem{FRVT-mask}
M.~L. Ngan, P.~J. Grother, K.~K. Hanaoka, et~al.
\newblock Ongoing {Face Recognition Vendor Test (FRVT) Part 6B}: Face
  recognition accuracy with face masks using post-{COVID}-19 algorithms.
\newblock 2021.

\bibitem{otto2017clustering}
C.~Otto, D.~Wang, and A.~K. Jain.
\newblock Clustering millions of faces by identity.
\newblock {\em TPAMI}, 2017.

\bibitem{VGGFace}
O.~M. Parkhi, A.~Vedaldi, and A.~Zisserman.
\newblock Deep face recognition.
\newblock In {\em BMVC}, 2015.

\bibitem{RegNet}
I.~Radosavovic, R.~P. Kosaraju, R.~Girshick, K.~He, and P.~Doll{\'a}r.
\newblock Designing network design spaces.
\newblock In {\em CVPR}, 2020.

\bibitem{ImageNet}
O.~Russakovsky, J.~Deng, H.~Su, J.~Krause, S.~Satheesh, S.~Ma, Z.~Huang,
  A.~Karpathy, A.~Khosla, M.~Bernstein, et~al.
\newblock Imagenet large scale visual recognition challenge.
\newblock In {\em IJCV}, 2015.

\bibitem{FaceNet}
F.~Schroff, D.~Kalenichenko, and J.~Philbin.
\newblock Facenet: A unified embedding for face recognition and clustering.
\newblock In {\em CVPR}, 2015.

\bibitem{CFP}
S.~Sengupta, J.-C. Chen, C.~Castillo, V.~M. Patel, R.~Chellappa, and D.~W.
  Jacobs.
\newblock Frontal to profile face verification in the wild.
\newblock In {\em WACV}, 2016.

\bibitem{PFE}
Y.~Shi and A.~K. Jain.
\newblock Probabilistic face embeddings.
\newblock In {\em ICCV}, 2019.

\bibitem{shi2020towards}
Y.~Shi, X.~Yu, K.~Sohn, M.~Chandraker, and A.~K. Jain.
\newblock Towards universal representation learning for deep face recognition.
\newblock In {\em CVPR}, 2020.

\bibitem{VGGNet}
K.~Simonyan and A.~Zisserman.
\newblock Very deep convolutional networks for large-scale image recognition.
\newblock {\em arXiv:1409.1556}, 2014.

\bibitem{sohn2016improved}
K.~Sohn.
\newblock Improved deep metric learning with multi-class n-pair loss objective.
\newblock In {\em NeurIPS}, 2016.

\bibitem{song2019occlusion}
L.~Song, D.~Gong, Z.~Li, C.~Liu, and W.~Liu.
\newblock Occlusion robust face recognition based on mask learning with
  pairwise differential siamese network.
\newblock In {\em ICCV}, 2019.

\bibitem{DeepID2}
Y.~Sun, Y.~Chen, X.~Wang, and X.~Tang.
\newblock Deep learning face representation by joint
  identification-verification.
\newblock In {\em NeurIPS}, 2014.

\bibitem{Deepid3}
Y.~Sun, D.~Liang, X.~Wang, and X.~Tang.
\newblock {DeepID3}: Face recognition with very deep neural networks.
\newblock {\em arXiv:1502.00873}, 2015.

\bibitem{DeepID}
Y.~Sun, X.~Wang, and X.~Tang.
\newblock Deep learning face representation from predicting 10,000 classes.
\newblock In {\em CVPR}, 2014.

\bibitem{DeepID2+}
Y.~Sun, X.~Wang, and X.~Tang.
\newblock Deeply learned face representations are sparse, selective, and
  robust.
\newblock In {\em CVPR}, 2015.

\bibitem{GoogleNet}
C.~Szegedy, W.~Liu, Y.~Jia, P.~Sermanet, S.~Reed, D.~Anguelov, D.~Erhan,
  V.~Vanhoucke, and A.~Rabinovich.
\newblock Going deeper with convolutions.
\newblock In {\em CVPR}, 2015.

\bibitem{DeepFace}
Y.~Taigman, M.~Yang, M.~Ranzato, and L.~Wolf.
\newblock Deepface: Closing the gap to human-level performance in face
  verification.
\newblock In {\em CVPR}, 2014.

\bibitem{taigman2015web}
Y.~Taigman, M.~Yang, M.~Ranzato, and L.~Wolf.
\newblock Web-scale training for face identification.
\newblock In {\em CVPR}, 2015.

\bibitem{tan2019efficientnet}
M.~Tan and Q.~Le.
\newblock Efficientnet: Rethinking model scaling for convolutional neural
  networks.
\newblock In {\em ICML}, 2019.

\bibitem{terhorst2021comprehensive}
P.~Terh{\"o}rst, J.~N. Kolf, M.~Huber, F.~Kirchbuchner, N.~Damer, A.~Morales,
  J.~Fierrez, and A.~Kuijper.
\newblock A comprehensive study on face recognition biases beyond demographics.
\newblock {\em arXiv:2103.01592}, 2021.

\bibitem{Flickr}
B.~Thomee, D.~A. Shamma, G.~Friedland, B.~Elizalde, K.~Ni, D.~Poland, D.~Borth,
  and L.-J. Li.
\newblock {YFCC100M}: The new data in multimedia research.
\newblock {\em Communications of the ACM}, 2016.

\bibitem{IMDB-Face}
F.~Wang, L.~Chen, C.~Li, S.~Huang, Y.~Chen, C.~Qian, and C.~Change~Loy.
\newblock The devil of face recognition is in the noise.
\newblock In {\em ECCV}, 2018.

\bibitem{AttentionNet}
F.~Wang, M.~Jiang, C.~Qian, S.~Yang, C.~Li, H.~Zhang, X.~Wang, and X.~Tang.
\newblock Residual attention network for image classification.
\newblock In {\em CVPR}, 2017.

\bibitem{AM-SoftMax}
F.~Wang, W.~Liu, H.~Liu, and J.~Cheng.
\newblock Additive margin softmax for face verification.
\newblock {\em IEEE SPL}, 2018.

\bibitem{NormFace}
F.~Wang, X.~Xiang, J.~Cheng, and A.~L. Yuille.
\newblock Normface: L2 hypersphere embedding for face verification.
\newblock In {\em ACM MM}, 2017.

\bibitem{wang2021pseudo}
G.~Wang, J.~Ma, Q.~Zhang, J.~Lu, and J.~Zhou.
\newblock Pseudo facial generation with extreme poses for face recognition.
\newblock In {\em CVPR}, pages 1994--2003, 2021.

\bibitem{wang2019decorrelated}
H.~Wang, D.~Gong, Z.~Li, and W.~Liu.
\newblock Decorrelated adversarial learning for age-invariant face recognition.
\newblock In {\em CVPR}, 2019.

\bibitem{CosFace}
H.~Wang, Y.~Wang, Z.~Zhou, X.~Ji, Z.~Li, D.~Gong, J.~Zhou, and W.~Liu.
\newblock Cos{F}ace: Large margin cosine loss for deep face recognition.
\newblock In {\em CVPR}, 2018.

\bibitem{wang2019mitigate}
M.~Wang and W.~Deng.
\newblock Mitigate bias in face recognition using skewness-aware reinforcement
  learning.
\newblock {\em arXiv:1911.10692}, 2019.

\bibitem{RFW}
M.~Wang, W.~Deng, J.~Hu, X.~Tao, and Y.~Huang.
\newblock Racial faces in the wild: Reducing racial bias by information
  maximization adaptation network.
\newblock In {\em CVPR}, 2019.

\bibitem{wang2020hierarchical}
Q.~Wang, T.~Wu, H.~Zheng, and G.~Guo.
\newblock Hierarchical pyramid diverse attention networks for face recognition.
\newblock In {\em CVPR}, 2020.

\bibitem{wang2019co}
X.~Wang, S.~Wang, J.~Wang, H.~Shi, and T.~Mei.
\newblock Co-mining: Deep face recognition with noisy labels.
\newblock In {\em ICCV}, 2019.

\bibitem{wang2019mis}
X.~Wang, S.~Zhang, S.~Wang, T.~Fu, H.~Shi, and T.~Mei.
\newblock Mis-classified vector guided softmax loss for face recognition.
\newblock In {\em AAAI}, 2020.

\bibitem{wang2020masked}
Z.~Wang, G.~Wang, B.~Huang, Z.~Xiong, Q.~Hong, H.~Wu, P.~Yi, K.~Jiang, N.~Wang,
  Y.~Pei, et~al.
\newblock Masked face recognition dataset and application.
\newblock {\em arXiv:2003.09093}, 2020.

\bibitem{wen2016discriminative}
Y.~Wen, K.~Zhang, Z.~Li, and Y.~Qiao.
\newblock A discriminative feature learning approach for deep face recognition.
\newblock In {\em ECCV}, 2016.

\bibitem{YTF}
L.~Wolf, T.~Hassner, and I.~Maoz.
\newblock Face recognition in unconstrained videos with matched background
  similarity.
\newblock In {\em CVPR}, 2011.

\bibitem{wu2018light}
X.~Wu, R.~He, Z.~Sun, and T.~Tan.
\newblock A light {CNN} for deep face representation with noisy labels.
\newblock {\em TIFS}, 2018.

\bibitem{noisy_student}
Q.~Xie, M.-T. Luong, E.~Hovy, and Q.~V. Le.
\newblock Self-training with noisy student improves {I}magenet classification.
\newblock In {\em CVPR}, 2020.

\bibitem{ResNeXt}
S.~Xie, R.~Girshick, P.~Doll{\'a}r, Z.~Tu, and K.~He.
\newblock Aggregated residual transformations for deep neural networks.
\newblock In {\em CVPR}, 2017.

\bibitem{xu2021consistent}
X.~Xu, Y.~Huang, P.~Shen, S.~Li, J.~Li, F.~Huang, Y.~Li, and Z.~Cui.
\newblock Consistent instance false positive improves fairness in face
  recognition.
\newblock In {\em CVPR}, 2021.

\bibitem{yalniz2019billion}
I.~Z. Yalniz, H.~J{\'e}gou, K.~Chen, M.~Paluri, and D.~Mahajan.
\newblock Billion-scale semi-supervised learning for image classification.
\newblock {\em arXiv:1905.00546}, 2019.

\bibitem{GCNV}
L.~Yang, D.~Chen, X.~Zhan, R.~Zhao, C.~C. Loy, and D.~Lin.
\newblock Learning to cluster faces via confidence and connectivity estimation.
\newblock In {\em CVPR}, 2020.

\bibitem{GCND}
L.~Yang, X.~Zhan, D.~Chen, J.~Yan, C.~C. Loy, and D.~Lin.
\newblock Learning to cluster faces on an affinity graph.
\newblock In {\em CVPR}, 2019.

\bibitem{CASIA-WebFace}
D.~Yi, Z.~Lei, S.~Liao, and S.~Z. Li.
\newblock Learning face representation from scratch.
\newblock {\em arXiv:1411.7923}, 2014.

\bibitem{zhang2018accelerated}
X.~Zhang, L.~Yang, J.~Yan, and D.~Lin.
\newblock Accelerated training for massive classification via dynamic class
  selection.
\newblock In {\em AAAI}, 2018.

\bibitem{zhang2019adacos}
X.~Zhang, R.~Zhao, Y.~Qiao, X.~Wang, and H.~Li.
\newblock Adacos: Adaptively scaling cosine logits for effectively learning
  deep face representations.
\newblock In {\em CVPR}, 2019.

\bibitem{zhang2019p2sgrad}
X.~Zhang, R.~Zhao, J.~Yan, M.~Gao, Y.~Qiao, X.~Wang, and H.~Li.
\newblock P2sgrad: Refined gradients for optimizing deep face models.
\newblock In {\em CVPR}, 2019.

\bibitem{MillionCelebs}
Y.~Zhang, W.~Deng, M.~Wang, J.~Hu, X.~Li, D.~Zhao, and D.~Wen.
\newblock Global-local {GCN}: Large-scale label noise cleansing for face
  recognition.
\newblock In {\em CVPR}, 2020.

\bibitem{zhao2018towards}
J.~Zhao, Y.~Cheng, Y.~Xu, L.~Xiong, J.~Li, F.~Zhao, K.~Jayashree, S.~Pranata,
  S.~Shen, J.~Xing, et~al.
\newblock Towards pose invariant face recognition in the wild.
\newblock In {\em CVPR}, 2018.

\bibitem{zhao2020towards}
J.~Zhao, S.~Yan, and J.~Feng.
\newblock Towards age-invariant face recognition.
\newblock {\em TPAMI}, 2020.

\bibitem{zhao2019regularface}
K.~Zhao, J.~Xu, and M.-M. Cheng.
\newblock Regular{F}ace: Deep face recognition via exclusive regularization.
\newblock In {\em CVPR}, 2019.

\bibitem{CPLFW}
T.~Zheng and W.~Deng.
\newblock Cross-pose {LFW}: A database for studying cross-pose face recognition
  in unconstrained environments.
\newblock Technical report, 2018.

\bibitem{CALFW}
T.~Zheng, W.~Deng, and J.~Hu.
\newblock Cross-age {LFW}: A database for studying cross-age face recognition
  in unconstrained environments.
\newblock {\em arXiv:1708.08197}, 2017.

\bibitem{zheng2021learning}
W.~Zheng, L.~Yan, F.-Y. Wang, and C.~Gou.
\newblock Learning from the web: Webly supervised meta-learning for masked face
  recognition.
\newblock In {\em CVPR Workshop}, 2021.

\bibitem{Zhong2019Unequal}
Y.~Zhong, W.~Deng, M.~Wang, J.~Hu, J.~Peng, X.~Tao, and Y.~Huang.
\newblock Unequal-training for deep face recognition with long-tailed noisy
  data.
\newblock In {\em CVPR}, 2019.

\bibitem{zhu2021masked}
Z.~Zhu, G.~Huang, J.~Deng, Y.~Ye, J.~Huang, X.~Chen, J.~Zhu, T.~Yang, J.~Guo,
  J.~Lu, et~al.
\newblock Masked face recognition challenge: The {WebFace260M} track report.
\newblock {\em arXiv preprint arXiv:2108.07189}, 2021.

\bibitem{WebFace260M}
Z.~Zhu, G.~Huang, J.~Deng, Y.~Ye, J.~Huang, X.~Chen, J.~Zhu, T.~Yang, J.~Lu,
  D.~Du, and J.~Zhou.
\newblock Web{F}ace260{M}: A benchmark unveiling the power of million-scale
  deep face recognition.
\newblock In {\em CVPR}, 2021.

\end{thebibliography}

\end{document}